\documentclass[sigconf]{acmart}

\AtBeginDocument{%
  }

\copyrightyear{2026}
\acmYear{2026}
\setcopyright{cc}
\setcctype{by}
\acmConference[WWW '26] {Proceedings of the ACM Web Conference 2026}{April 13--17, 2026}{Dubai, United Arab Emirates.}
\acmBooktitle{Proceedings of the ACM Web Conference 2026 (WWW '26), April 13--17, 2026, Dubai, United Arab Emirates}
\acmISBN{979-8-4007-2307-0/2026/04}
\acmDOI{10.1145/xxxxxxxxxxxx}

\usepackage{algorithm}
\usepackage{algorithmic}
\usepackage[dvipsnames]{xcolor}
\usepackage{color}
\usepackage{balance}
\usepackage{colortbl}

\usepackage{amsmath,amsfonts,bm}

\def\eqref#1{equation~\ref{#1}}

\def\1{\bm{1}}

\def\rvx{{\mathbf{x}}}

\def\mA{{\bm{A}}}
\def\mB{{\bm{B}}}

\def\mE{{\bm{E}}}

\def\mH{{\bm{H}}}
\def\mI{{\bm{I}}}

\def\mK{{\bm{K}}}

\def\mM{{\bm{M}}}
\def\mN{{\bm{N}}}

\def\mP{{\bm{P}}}
\def\mQ{{\bm{Q}}}

\def\mS{{\bm{S}}}

\def\mV{{\bm{V}}}
\def\mW{{\bm{W}}}
\def\mX{{\bm{X}}}
\def\mY{{\bm{Y}}}
\def\mZ{{\bm{Z}}}

\DeclareMathAlphabet{\mathsfit}{\encodingdefault}{\sfdefault}{m}{sl}
\SetMathAlphabet{\mathsfit}{bold}{\encodingdefault}{\sfdefault}{bx}{n}

\def\gG{{\mathcal{G}}}

\usepackage{pifont}

\usepackage{xspace}
\makeatletter
\DeclareRobustCommand\onedot{\futurelet\@let@token\@onedot}
\def\@onedot{\ifx\@let@token.\else.\null\fi\xspace}

\def\eg{\emph{e.g}\onedot} 
\def\ie{\emph{i.e}\onedot}

\def\etal{\emph{et al}\onedot}
\makeatother

\usepackage{overpic}

\usepackage{color}
\definecolor{cvprblue}{rgb}{0.21,0.49,0.74}

\usepackage{enumitem}
\usepackage{multirow}
\usepackage[dvipsnames]{xcolor}
\usepackage{listings}

\usepackage{subcaption}
\usepackage{xcolor}
\newtheorem{definition}{Definition}
\newtheorem{theorem}{Theorem}
\usepackage{newfloat}
\usepackage{listings}
\usepackage{bibentry}
\begin{document}

\title{GraphTARIF: Linear Graph Transformer with Augmented Rank and Improved Focus}






\author{Zhaolin Hu}
\orcid{0009-0002-7928-9638}
\affiliation{%
\institution{Zhejiang University}
\city{Hangzhou}
\state{Zhejiang}
\country{China}
}
\email{12321165@zju.edu.cn}

\author{Kun Li}
\orcid{0000-0001-5083-2145}
\affiliation{%
\institution{United Arab Emirates University}
\city{Al Ain}
\country{United Arab Emirates}
}
\email{kunli.hfut@gmail.com}

\author{Hehe Fan}
\authornote{Corresponding author.}
\orcid{0000-0001-9572-2345}
\affiliation{%
\institution{Zhejiang University}
\city{Hangzhou}
\state{Zhejiang}
\country{China}
}
\email{hehefan@zju.edu.cn}

\author{Yi Yang}
\orcid{0000-0002-0512-880X}
\affiliation{%
\institution{Zhejiang University}
\city{Hangzhou}
\state{Zhejiang}
\country{China} 
}
\email{yangyics@zju.edu.cn}
\renewcommand{\shortauthors}{Zhaolin Hu, Kun Li, Hehe Fan, and Yi Yang}

\begin{abstract}
  Linear attention mechanisms have emerged as efficient alternatives to full self-attention in Graph Transformers, offering linear time complexity. However, existing linear attention models often suffer from a significant drop in expressiveness due to low-rank projection structures and overly uniform attention distributions. We theoretically prove that these properties reduce the class separability of node representations, limiting the model’s classification ability. To address this, we propose a novel hybrid framework that enhances both the rank and focus of attention. Specifically, we enhance linear attention by attaching a gated local graph network branch to the value matrix, thereby increasing the rank of the resulting attention map. Furthermore, to alleviate the excessive smoothing effect inherent in linear attention, we introduce a learnable log-power function into the attention scores to reduce entropy and sharpen focus. We theoretically show that this function decreases entropy in the attention distribution, enhancing the separability of learned embeddings. Extensive experiments on both homophilic and heterophilic graph benchmarks demonstrate that our method achieves competitive performance while preserving the scalability of linear attention.
\end{abstract}




\maketitle

\section{Introduction}
Graph Neural Networks (GNNs)~\cite{scarselli2008graph, kipf2016semi, hamilton2017inductive, velivckovic2017graph,jian2025reaction} have emerged as a powerful paradigm for analyzing graph-structured data in a variety of Web-related domains, including social networks~\cite{chen2017supervised}, recommender systems~\cite{ying2018graph}, communication networks~\cite{lu2024gcepnet}, and anomaly detection ~\cite{dou2020enhancing}. By leveraging message passing across graph neighborhoods, GNNs effectively capture local structural patterns and have achieved remarkable success in node classification, link prediction, and graph-level inference tasks. Nevertheless, their reliance on strictly local aggregation makes it inherently difficult to model long-range dependencies across distant nodes~\cite{alon2020bottleneck}, thereby restricting expressiveness and limiting their ability to approximate complex relational functions in large-scale Web graphs~\cite{xu2018powerful}.

Inspired by recent advances in Transformer-based models in language and vision domains~\cite{vaswani2017attention, dosovitskiy2020image,fan2022point,li2023vigt,chen2025prompt}, Graph Transformers (GTs) ~\cite{ying2021transformers, ma2023graph} have gained increasing popularity in recent years. Unlike GNNs, GTs allow each node to directly attend to every other node in the graph, effectively creating a fully connected graph to mitigate locality bias and capture long-range interactions between distant nodes. The global self-attention (SA) mechanism can enhance the expressivity of the model and has demonstrated considerable success, especially in graph-level tasks such as molecular property prediction~\cite{ma2023graph}.

Nevertheless, the quadratic complexity of full self-attention ($O(N^2)$ for $N$ nodes) poses a significant bottleneck for scaling Graph Transformers to node-level tasks on large-scale Web graphs, such as social networks or e-commerce interaction graphs, which often contain millions of nodes and edges. To address this challenge, recent works introduce linear self-attention (LSA) mechanisms~\cite{wu2023sgformer, wu2022nodeformer, deng2024polynormer}, which approximate full softmax attention through kernel feature mappings or low-rank projections. This design reduces both memory and runtime costs from $O(N^2)$ to $O(N + |E|)$, enabling GTs to process much larger graphs more efficiently. Despite these computational benefits, empirical studies show that LSA-based Graph Transformers underperform  even classical GNNs across a variety of benchmarks~\cite{luo2024classic, zhou2024rethinking}, raising concerns about their effectiveness in Web applications where both scalability and accuracy are critical.

To study the limitations of linear GT and improve its performance, we first discover the loss of rank in overly uniform (high-entropy) attention distribution issues of LSA compared to full SA through empirical studies. Intuitively, a low-rank LSA matrix restricts the model to express only a limited subset of distinct attention patterns, potentially collapsing the representations of different nodes or even different classes into similar embeddings. At the same time, high-entropy attention distributions lead to nearly uniform weights across nodes, preventing the model from sharply focusing on the most informative neighbors. To analyze the above two issues, we provide a theoretical analysis showing that the low-rank property of LSA matrices significantly reduces the differences between inter-class node representations, which has negative impact on node distinguishability. In addition, we conduct extensive experiments to demonstrate that high-entropy attention distributions hurt the performance of LSA by overly smoothing node representations.

To address these issues, we propose a novel linear Graph Transformer architecture (GraphTARIF) designed to increase the rank of attention scores and lower their entropy. Our method integrates a local module into linear attention to enhance expressiveness and introduces a learnable log-power function to sharpen attention distributions while maintaining training stability. The resulting model is a High-Rank Low-Entropy linear graph transformer, tailored explicitly for node-level tasks. We achieve competitive results in a wide range of homophilic and heterophilic graph learning tasks. Our contributions are summarized below:
\begin{itemize}
\item  We theoretically prove that the low-rank and high-entropy properties of linear self-attention significantly reduce inter-class distinguishability, providing a principled explanation for the underperformance of linear Graph Transformers on node-level tasks.   
\item  We introduce GraphTARIF, a linear Graph Transformer that integrates a local enhancement module and a learnable log-power function to simultaneously improve attention rank and reduce entropy, striking a balance between scalability and expressiveness.  
\item  We evaluate GraphTARIF across a wide spectrum of Web-related datasets, \ie, e-commerce, social networks, crowdsourcing platforms, and Wikipedia-based graphs,  and it consistently outperforms existing baselines.  
\end{itemize}

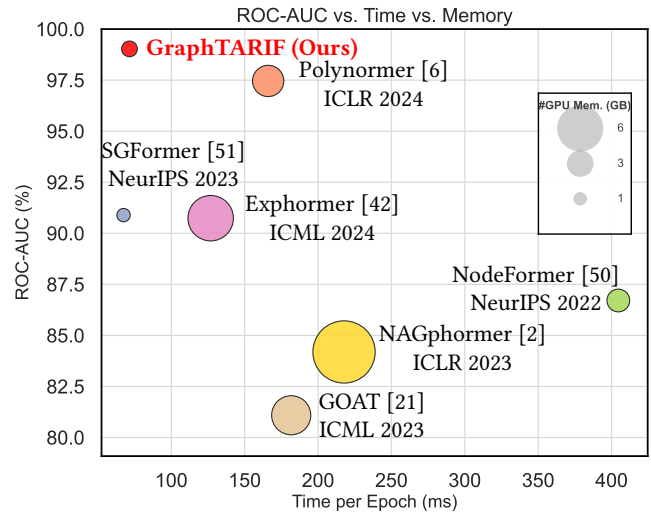
\begin{figure}[t!]
\centering
\begin{overpic}[width=1.0\linewidth,tics=10]{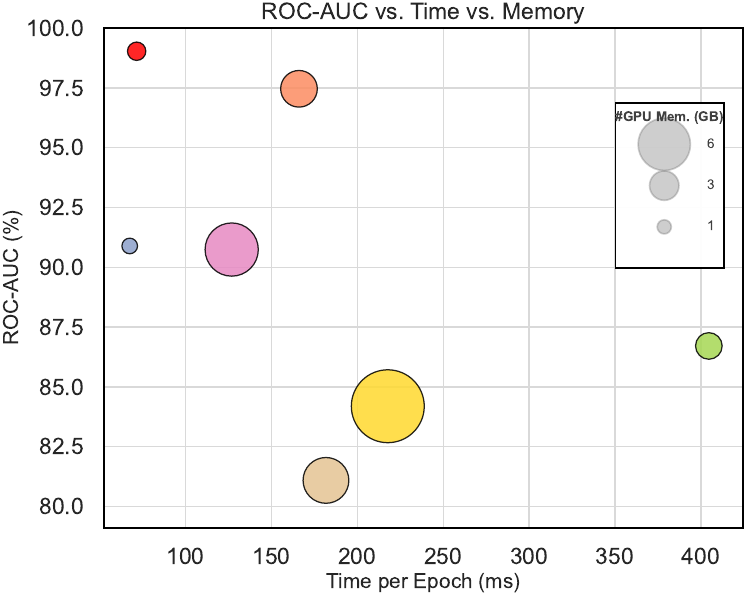} 
\put(21,72){\textcolor{red}{\textbf{GraphTARIF (Ours)}}}
\put(44,64){
\shortstack[c]{Polynormer~\cite{deng2024polynormer}\\ICLR 2024}}
\put(36.5,43){\shortstack[c]{Exphormer~\cite{shirzad2023exphormer}\\ ICML 2024}}
\put(14,51.5){\shortstack[c]{SGFormer~\cite{wu2023sgformer} \\ NeurIPS 2023}}
\put(69,32){\shortstack[c]{NodeFormer~\cite{wu2022nodeformer} \\ NeurIPS 2022}}
\put(57.5,22.5){\shortstack[c]{NAGphormer~\cite{chen2023nagphormer}\\ ICLR 2023}}
\put(48.0,12){\shortstack[c]{GOAT~\cite{kong2023goat} \\ ICML 2023}}
\end{overpic}
\caption{Comparisons among Graph Transformers(GT) in terms of accuracy, runtime, and GPU memory usage on Minesweeper. 
The proposed GraphTARIF achieves consistently superior accuracy with shorter runtime, demonstrating both effectiveness and efficiency.}
\label{fig:intro11}
\end{figure}

\section{Related Work}
\textbf{Graph Neural Networks.} Graph Neural Networks (GNNs) have emerged as powerful methods for node classification tasks, primarily relying on local neighborhood aggregation through message passing. For example, Graph Convolutional Networks (GCNs)~\cite{kipf2016semi} aggregate neighborhood information by leveraging a normalized adjacency matrix to propagate features. GraphSAGE~\cite{hamilton2017inductive} generalizes this approach by sampling and aggregating neighboring nodes, enabling inductive learning. Graph Attention Networks (GATs)~\cite{velivckovic2017graph} introduce attention mechanisms at the edge level, adaptively weighting neighbors based on their importance. Despite their success, these methods inherently suffer from local heterophily~\cite{luan2024heterophilic, luan2024graph, lu2024flexible, zheng2024missing, zheng2025let}, oversmoothing~\cite{li2018deeper, luan2019break}, and oversquashing~\cite{alon2020bottleneck, topping2021understanding}, limiting their capability to represent complex node interactions across long-range distances.

\noindent \textbf{Graph Transformer.} Graph Transformers (GTs) employ global self-attention mechanisms, which allow direct interactions between distant nodes, thus mitigating locality bias. Early graph transformer models, such as Graph-BERT~\cite{zhang2020graph} and Graph Transformer~\cite{dwivedi2020generalization}, demonstrated the potential of self-attention for capturing long-range dependencies. However, the quadratic complexity of standard Transformer attention~\cite{vaswani2017attention} is computationally prohibitive for large-scale graphs. Recent works have addressed this through linear attention approximations, such as Nodeformer~\cite{wu2022nodeformer} and SGformer~\cite{wu2023sgformer}, significantly improving computational efficiency. Nonetheless, studies have highlighted limitations of linear attention, specifically low-rank attention matrices~\cite{fan2024breaking} and uniformity (high-entropy )~\cite {han2023flatten, meng2025polaformer} of attention score distributions, leading to suboptimal node classification performance. However, these problems are underexplored for linear graph transformers.

In this paper, we empirically demonstrate and theoretically analyze the negative impacts of these limitations. Then, we propose a novel Graph Transformer architecture to enhance attention rank and reduce attention entropy.

\section{Preliminaries}

Define a graph as $\gG = (\mathcal{V}, \mathcal{E}, \mX, \mY) $, where $ \mathcal{V} $ is the set of nodes with $ |\mathcal{V}| = n $, and $ \mathcal{E} \subseteq \mathcal{V} \times \mathcal{V} $ denotes the set of edges. Let $ \mX \in \mathbb{R}^{n \times d} $ be the node feature matrix, where $d$ is the feature dimension, and $\mY \in \mathbb{R}^{n \times C}$ the one-hot encoded label matrix with $C$ classes. The adjacency matrix of $\gG$ is denoted by $\mA \in \mathbb{R}^{n \times n} $.

\subsection{Graph Neural Networks}
In the message passing framework~\cite{gilmer2017neural}, Graph Neural Networks (GNNs) update node representations by aggregating information from their neighbors. A typical GNN layer updates the representation of node $i$ as:
\begin{equation}
h_i^{(l+1)} = f^{(l)}\left(h_i^{(l)}, \text{AGG}^{(l)}\left(\{ h_j^{(l)} : j \in \mathcal{N}_i \}\right)\right),
\end{equation}
where $h_i^{(l)}$ is the embedding of node $i$ at the $l$-th layer, $\mathcal{N}_i$ denotes the neighbors of node $i$, $f^{(l)}$ is the activation function, and $\text{AGG}^{(l)}(\cdot)$ is an aggregation function such as sum, mean, or max pooling~\cite{xu2018powerful}.

\subsection{Graph Transformers}

\subsubsection{Standard Self-Attention.}
Transformers~\cite{vaswani2017attention} use self-attention to capture global dependencies between all pairs of nodes. The attention mechanism is defined as:
\begin{equation}
\text{Attention}(\mQ, \mK, \mV) = \text{softmax}\left(\frac{\mQ\mK^\top}{\sqrt{d_k}}\right)\mV,
\end{equation}
\begin{equation}
\mQ=\mH \mW_Q, \mK=\mH \mW_K, \mV=\mH \mW_V,
\end{equation}
where $\mQ, \mK, \mV$ are query, key, and value matrices obtained via learned projections of input node features $H$, and $d_k$ is the dimension of queries/keys. Although powerful, this formulation has $\mathcal{O}(N^2)$ time and space complexity, making it unsuitable for large-scale graphs.

\subsubsection{Linear Attention.}
To address the challenge of scalability, linear attention mechanisms approximate softmax attention with a kernel function that allows rearranging the computation to linear complexity~\cite{katharopoulos2020transformers}:
\begin{equation} 
\text{LinearAttention}(\mQ, \mK, \mV) = \text{Sim}(\mQ, \mK) \mV = \phi(\mQ) \left( \phi(\mK)^\top \mV \right),
\end{equation}
where $\phi(\cdot)$ is a kernel feature map, such as positive random features or ReLU-based transformations~\cite{katharopoulos2020transformers}. This reduces the complexity from $\mathcal{O}(N^2)$ to $\mathcal{O}(N)$, enabling applications to large graphs. However, in the following section, we will show that such an approximation often leads to \textit{low-rank} attention matrices and \textit{high-entropy} attention distributions, which can limit expressiveness and harm model performance on node classification tasks.

\subsection{Entropy}

Entropy measures the uncertainty or dispersion of a sequence. Lower entropy indicates a more concentrated distribution, which often implies clearer semantic focus. In contrast, high entropy corresponds to smoother distributions that may dilute important signals. In this work, we adopt the definition of entropy for nonnegative sequences as proposed in prior work~\cite{meng2025polaformer}.
\begin{definition}[Positive Sequence Entropy (PSE)~\cite{meng2025polaformer}]
Let a sequence $\mathbf{x} = (x_1, \dots, x_N)$ with $x_i \geq 0$, and $s = \sum_{i=1}^N x_i > 0$. Then the entropy of this positive sequence is defined by:
\begin{equation}
\mathrm{PSE}(\mathbf{x}) = -\sum_{i=1}^N \frac{x_i}{s} \log \left( \frac{x_i}{s} \right).
\end{equation}
\end{definition}

\section{Motivation}
In this section, we empirically and theoretically analyze the inherent limitations of linear attention in the context of node classification.

We build a minimal vanilla Graph Transformer architecture and evaluate it on three datasets: two homophilic graphs (WikiCS~\cite{mernyei2020wiki} and CS~\cite{shchur2018pitfalls}) and one heterophilic graph (Toloker~\cite{platonov2023critical}). To incorporate structural information, we adopt Random Walk Structural Encoding~\cite{dwivedi2021graph} (RWSE) as the positional encoding method. 

\begin{figure}[t!]
\centering
\begin{subfigure}{0.31\textwidth}
\centering
\includegraphics[width=\linewidth]{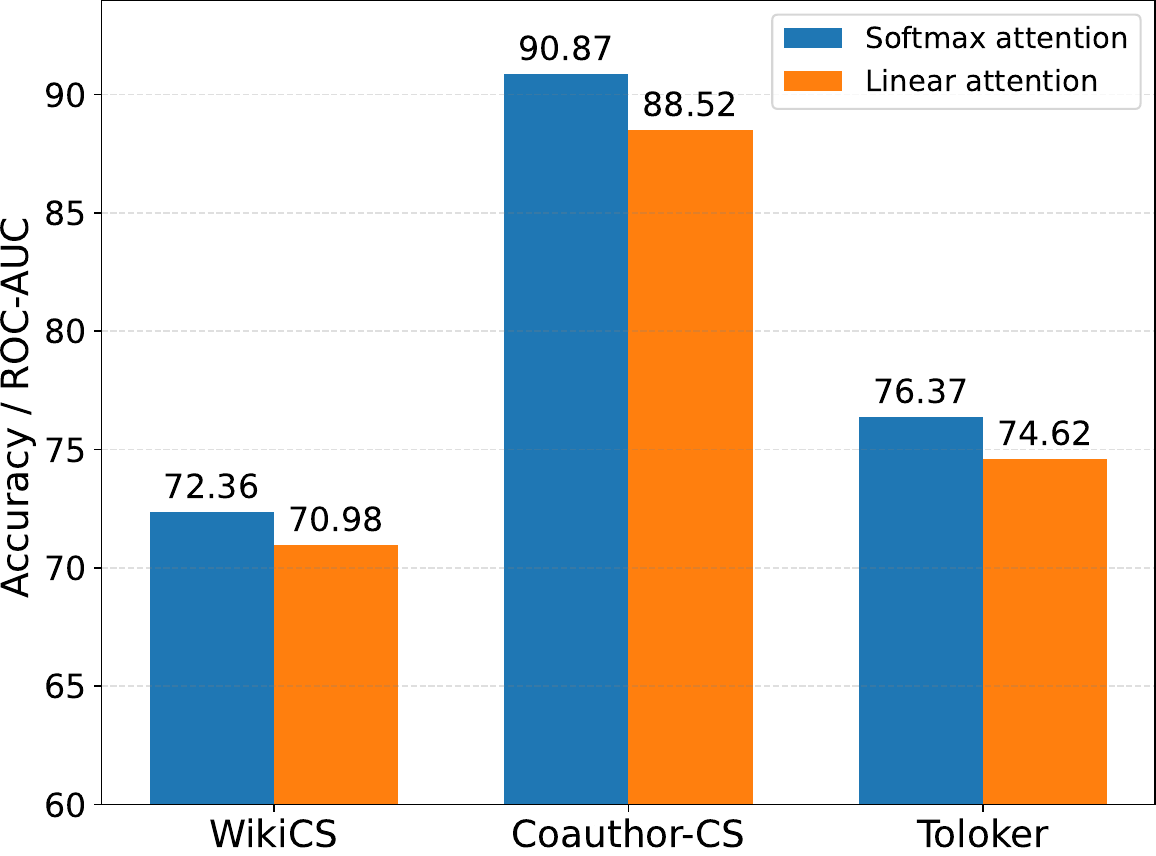}
\caption{Node classification accuracy}
\end{subfigure}%
\hspace{0.5em}
\begin{subfigure}{0.48\textwidth}
\centering
\includegraphics[width=\linewidth]{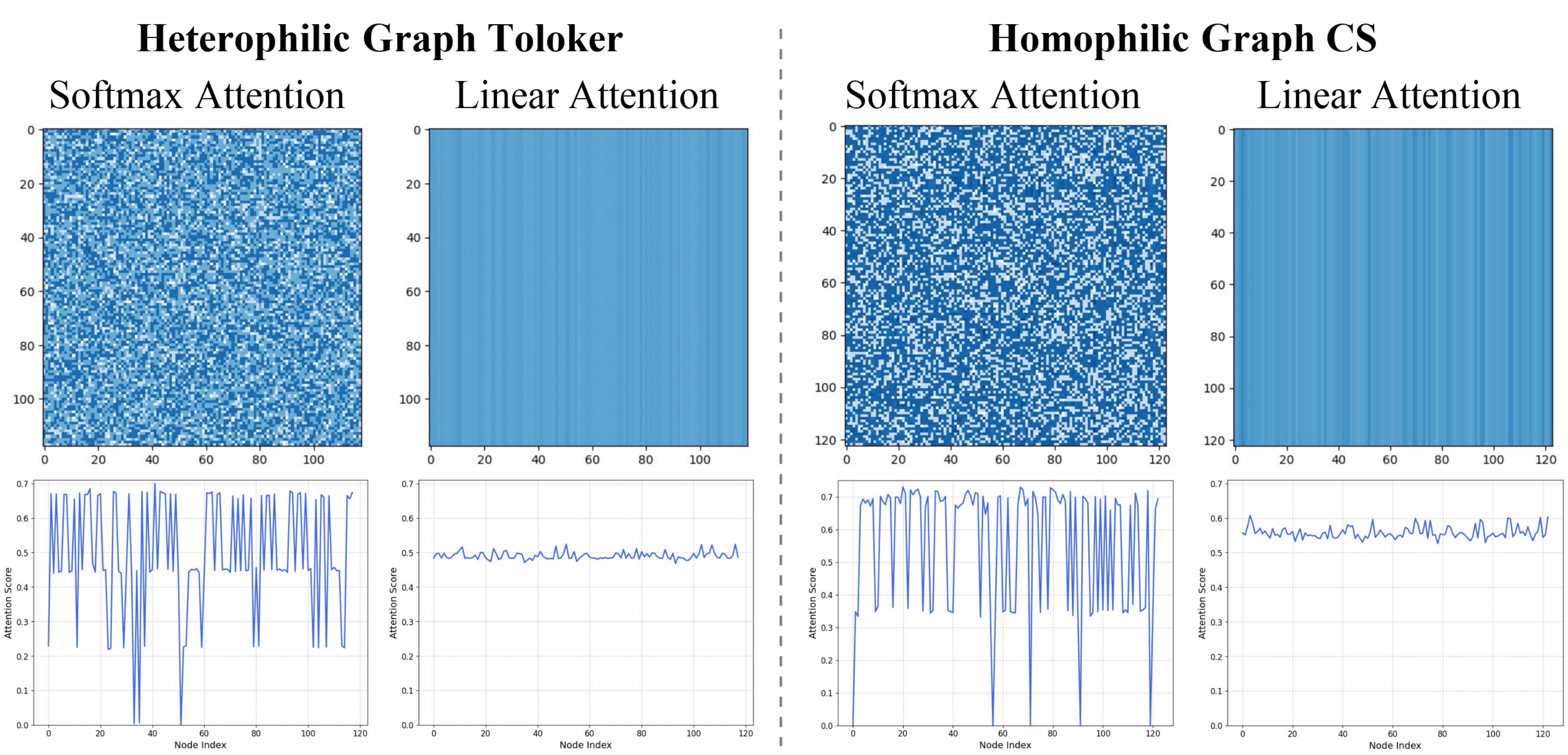}
\caption{Visualization of normalized attention matrices}
\end{subfigure}
\caption{(a) Node classification performance on homophilic graphs (WikiCS, CS) and a heterophilic graph (Toloker) shows that softmax attention consistently outperforms linear attention.
(b) Visualization of normalized attention matrices for 120 evenly-sampled nodes reveals that linear attention produces low-rank, high-entropy attention distributions across both homophilic and heterophilic graphs.}
\label{fig:intro}
\end{figure}

As shown in Fig.~\ref{fig:intro} (a), across all three datasets, the variant with softmax attention significantly outperforms its linear counterpart. This trend holds true across both homophilic and heterophilic graphs. 
Similar to its behavior in textual and visual transformers, linear attention also underperforms compared to softmax attention in graph-based tasks.

To better understand the underlying reasons, we visualize the attention score matrices for both attention types. For clarity, we uniformly sample approximately 120 nodes from the graph and normalize the attention vectors row-wise for visualization. As shown in Fig.~\ref{fig:intro} (b), we observe that:
\begin{itemize}
\item Linear attention consistently produces \textbf{low-rank} attention matrices—much lower than the number of nodes $N$;
\item The attention weights are \textbf{smooth and uniform}, indicating high entropy compared to softmax attention.
\end{itemize}
These phenomena are observed in both homophilic and heterophilic graphs, suggesting that the issues are fundamental rather than dataset-specific.

\begin{figure*}[t!]
\centering
\includegraphics[width=0.80\textwidth]{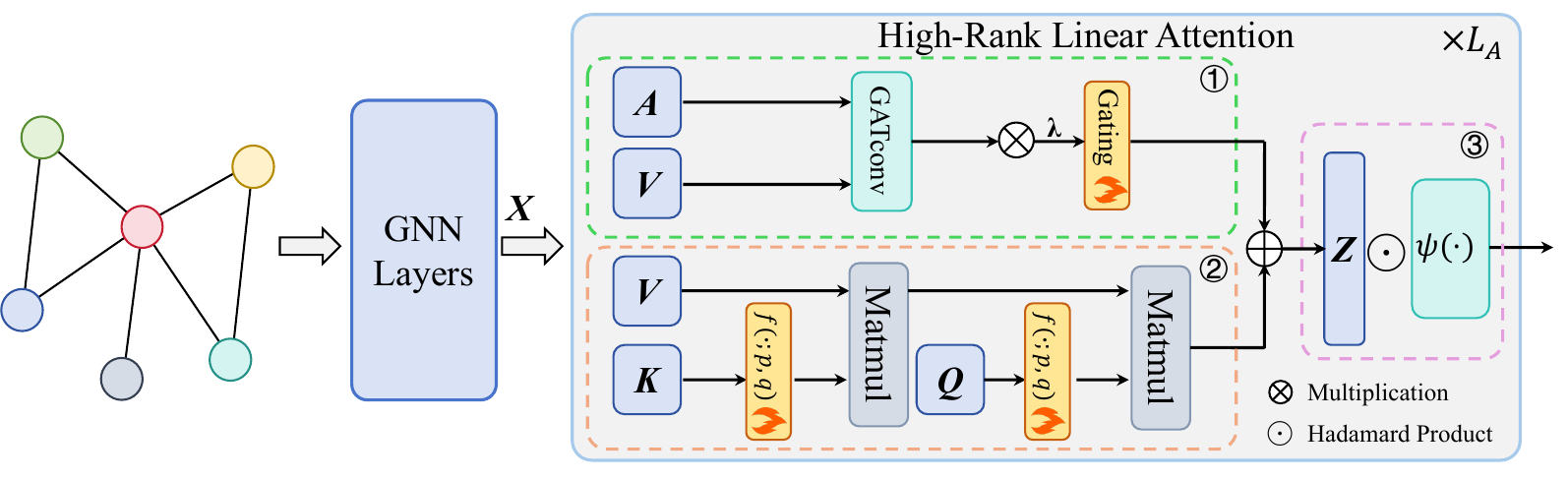} 
\caption{
The overall framework of GraphTARIF. It mainly consists of GNN layers and high-rank linear attention. \ding{172} is the gated local attention that mitigates the low-rank limitation of linear attention. \ding{173} is a learnable log-power function $f(\cdot; p, q)$ that sharpens the attention distribution and reduces entropy, and \ding{174} is a node-wise post-modulation module that produces clearer and more discriminative node representations. 
Here, $\mQ, \mK, \mV$ denote the standard query, key, and value projections of node features $\mX$, 
$\mA$ is the graph adjacency matrix, and $\psi(\cdot)$ denotes a simple linear projection applied for node-wise post-modulation. 
}
\label{fig:main}
\end{figure*}

To understand this behavior, let us examine the structural limitation of linear attention. Suppose $\phi(\mQ), \phi(\mK) \in \mathbb{R}^{N \times d}$. The attention score matrix is computed as:
\begin{equation}
\mM = \phi(\mQ) \phi(\mK)^\top \quad \text{with} \quad \operatorname{rank}(\mM) \leq \min\{N, d\}.
\end{equation}
Since $d$ is typically small in practice, and $\mN$ can range from tens to hundreds of thousands in large-scale graph datasets, the resulting attention matrix $\mM$ suffers from a severe rank bottleneck. This limitation significantly restricts the model's capacity to capture diverse and fine-grained interactions among nodes.

Furthermore, Zhang~\etal~\cite{zhang2024hedgehog} demonstrate that linear attention tends to produce smooth and high-entropy attention distributions. This issue becomes even more pronounced in graph-based tasks with a large number of nodes, where the attention weights are excessively dispersed. Such diffuse attention exacerbates the rank deficiency of the attention matrix, ultimately hindering the model’s expressiveness.

We now present two theoretical results that formally characterize how low-rank attention impacts the model's ability to preserve class separability.

In node classification, the objective is to learn node representations such that nodes belonging to the same class are embedded closely together, while those from different classes are well-separated. A fundamental metric for evaluating this separability is the inter-class scatter matrix $\mS_B$, defined in its unweighted form as:
\begin{equation}
\mS_B := \frac{1}{K} \sum_{k=1}^{K} (\boldsymbol{\mu}_k - \boldsymbol{\mu})(\boldsymbol{\mu}_k - \boldsymbol{\mu})^\top,
\end{equation}
where $K$ is the number of classes, $\boldsymbol{\mu}_k$ denotes the mean embedding of nodes in class $k$, and $\boldsymbol{\mu}$ is the global mean of all node embeddings. The trace $\mathrm{tr}(\mS_B)$ quantifies the total variance among class centroids. In this work, we adopt the trace of the inter-class scatter matrix, $ \mathrm{tr}(\mS_B) $, as a quantitative measure of inter-class variance in node classification tasks. Specifically, $\mathrm{tr}(\mS_B)$ quantifies the total variance among class centroids. A larger $ \mathrm{tr}(\mS_B) $ indicates greater inter-class separation, which typically implies stronger discriminative power and higher classification accuracy.

\begin{theorem}\label{theorem:1}
Let $\mX \in \mathbb{R}^{n \times d} $ be the node feature matrix and $\mM \in \mathbb{R}^{n \times n} $ an attention matrix applied to transform the embeddings. Suppose that the rows of $\mX$ are drawn from a Gaussian mixture model. The expected inter-class variance after applying the attention transformation, measured by the trace of the between-class scatter matrix, is upper bounded by a term that scales linearly with the rank of $\mM$:
\begin{equation}
\mathbb{E}[\mathrm{tr}(\mS_B(\mM \mX))] \leq C \cdot r,
\end{equation}
where $r$ is the rank of the attention matrix $\mM$, and $C$ is a constant that depends on the underlying graph data and feature distribution.
\end{theorem}

The proof is provided in Appendix~\ref{sec:app_proof1}. This theorem suggests that severely low-rank attention mechanisms may limit the ability to preserve class-discriminative information.

\begin{theorem}\label{theorem:2}
Let $\mM_1$ be an attention matrix, and let $\mM_2$ be obtained by applying a low-rank and smoothing transformation to $\mM_1$, \eg, $ \mM_2 = \mP \mM_1 $, where $ \mP \in \mathbb{R}^{n \times n} $ is a low-rank and smoothing operator. Then,
\begin{equation}
\mathbb{E}[\mathrm{tr}(\mS_B(\mM_2 \mX))] < \mathbb{E}[\mathrm{tr}(\mS_B(\mM_1 \mX))].
\end{equation}
\end{theorem}

The proof is given in Appendix~\ref{sec:app_proof2}. This result formalizes the intuition that applying a smoother and lower-rank transformation to the attention matrix reduces its ability to preserve discriminative structure.


These results support our core hypothesis: the low-rank, overly smooth nature of linear attention limits its ability to preserve inter-class variation in node representations, explaining its degraded performance in node classification. 
Motivated by this, the remainder of the paper introduces architectural modifications designed to increase the effective rank of the attention matrix and sharpen attention distributions, thereby enhancing class separability and improving classification accuracy.

\section{Methodology}
Based on the observations and theoretical analysis presented in the previous section, we propose a novel Graph Transformer architecture specifically designed to address the low-rank and high-entropy limitations of linear attention. 
As illustrated in Fig.~\ref{fig:main}, our method comprises two core components: a High-Rank Linear Attention mechanism and a learnable log-power function aimed at reducing.

\subsection{High-Rank Linear Attention}
To address the inherent low-rank limitation of standard linear attention mechanisms, we propose a simple yet effective strategy to enhance the rank of the attention output. In its conventional form, linear attention is computed as:
\begin{equation}
\mZ = \phi(\mQ) \left(\phi(\mK)^\top \mV\right).
\end{equation}

Inspired by the previous works~\cite{han2023flatten}, we introduce an additional complementary branch to enrich the representation by injecting higher-rank local context. Specifically, we modify the attention output as follows:
\begin{equation}
\mZ = \phi(\mQ) \left(\phi(\mK)^\top \mV\right) + \text{GAT}(\mV), 
\end{equation}
where $\text{GAT}(\mV)$ denotes the local features generated by a Graph Attention Network (GAT). This modification is equivalent to implicitly augmenting the attention score matrix from $=\phi(\mQ)\phi(\mK)^\top$ to $\mM_{\text{eq}}:=\phi(\mQ)\phi(\mK)^\top + M_{\text{GAT}}$,
in which $\mM_{\text{GAT}}$ corresponds to the attention matrix produced by the GAT module, and \( \mM_{\text{eq}} \) refers to the resulting equivalent attention matrix. Since GAT assigns attention weights based on the local neighborhood defined by the graph structure, $\mM_{\text{GAT}}$ is typically sparse, with each row normalized over the corresponding node’s immediate neighbors. Despite this sparsity, $\mM_{\text{GAT}}$ can, in principle, attain a rank as high as $n - 1$. This greater rank capacity enables $\mM_{\text{eq}}$ to capture more expressive and diverse node-level interactions, thus enhancing the model’s representational power.

However, we observe that simply incorporating GAT attention does not consistently improve performance~\cite{deng2024polynormer}. We attribute this to the numerical dominance of GAT-derived attention scores in node classification tasks. Unlike vision transformers, where attention score matrices typically have only a few hundred dimensions, attention matrices in graph-based models often span tens of thousands of dimensions due to the large number of nodes. Moreover, since GAT assigns non-zero attention weights only to a node's local neighbors, the resulting attention values tend to be significantly larger in magnitude than those from global linear attention. When combined without proper regulation, this disparity can lead to the suppression of global signals, thereby undermining the benefits of global context modeling.

To mitigate the dominance of local GAT attention in the hybrid formulation, we introduce a principled gating mechanism referred to as the Gating Mechanism that modulates its influence in a soft and learnable manner. Specifically, we incorporate a scalar gate into the GAT branch, resulting in the following enhanced formulation:
\begin{equation}
\mZ = \phi(\mQ) \left(\phi(\mK)^\top \mV\right) + \lambda \cdot \sigma(a) \cdot \text{GAT}(\mV),
\end{equation}
where $\lambda > 0$ is a global scaling hyperparameter, and $\sigma(a)$ is a gating function (\eg, sigmoid) applied to a learnable scalar $a$. This design introduces a trainable regulator that balances global and local contributions during training.

Overall, the proposed hybrid attention mechanism retains the computational efficiency of linear attention while incorporating an interpretable and controllable local inductive bias. This formulation effectively increases the rank of the attention score matrix and enhances the model's representational capacity. Moreover, based on our experiments and prior studies~\cite{xing2024less}, global graph attention often allocates excessive weights to distant nodes, which leads to overly smooth output representations and weakens the preservation of node-specific information.

To alleviate this issue, we adopt a node-wise post-modulation strategy following the global attention propagation. Specifically, inspired by~\cite{fan2024breaking}, we reweight the output of the attention mechanism using a transformed version of the original node features, denoted by $\psi(\mX)$. The complete form of our module is:
\begin{equation}
\bar{\mZ} = \psi(\mX) \odot \mZ.
\label{post-m}
\end{equation}
We use a simple linear projection to serve as the function $\psi(\cdot)$.
This modulation adjusts each node’s output based on its own features, helping to retain node-specific information while incorporating contextual signals. Previous works~\cite{fan2024breaking} have shown that this operation can increase the rank of the output representations. We theoretically demonstrate that it also reduces the entropy of the output representations, resulting in sharper and more discriminative embeddings, as detailed in  Appendix~\ref{sec:app_proof4}. 

In summary, we propose a simple yet effective architecture that increases the rank of the attention scores matrix and reduces the entropy of the output representations, thereby enhancing the model’s capacity to produce discriminative node embeddings. Detailed visualizations can be found in Figs.~\ref{fig:lambda-xrank}, and ~\ref{fig:lambda-tsne} of Appendix~\ref{sec:additional_results}.


\subsection{Learnable Log-Power Sharpening Function}

\begin{figure}[t!]
\centering
\begin{subfigure}{0.44\linewidth}
\centering
\includegraphics[width=\linewidth]{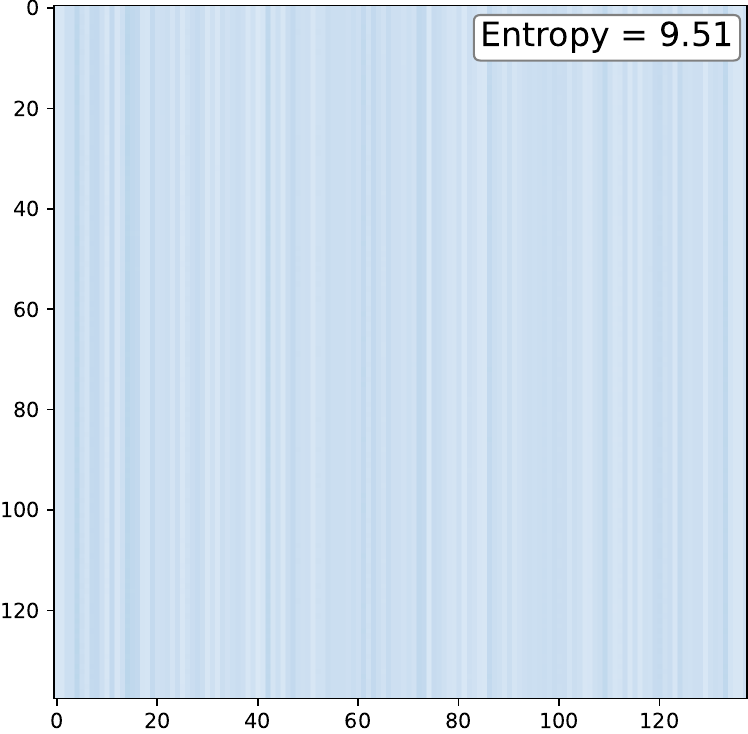}
\caption{Linear Attention}
\end{subfigure}
\hfill
\begin{subfigure}{0.44\linewidth}
\centering
\includegraphics[width=\linewidth]{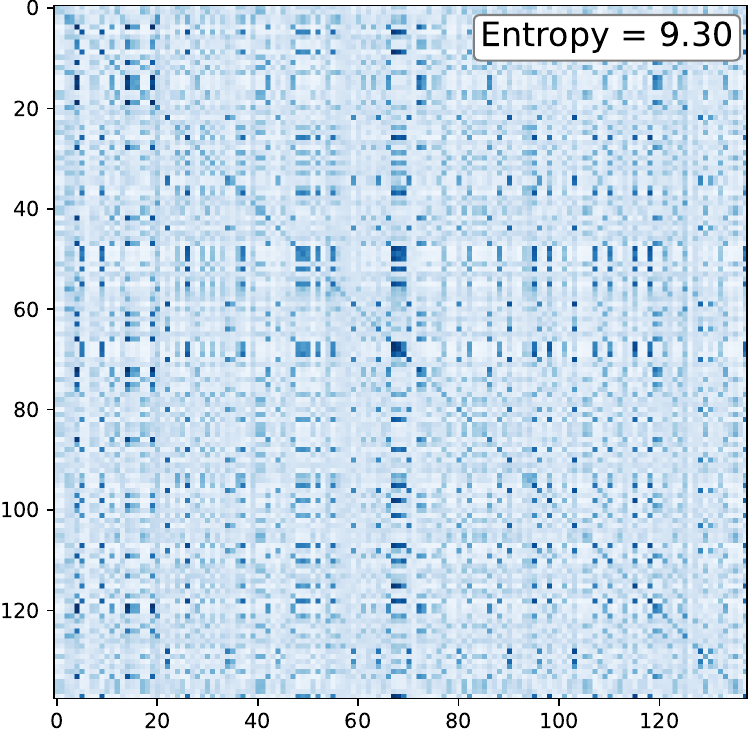}
\caption{GraphTARIF (Ours)}
\end{subfigure}
\caption{(a) Linear attention yields high-entropy attention maps with smooth, overly uniform outputs. (b) After applying the Learnable Log-Power Functions, the attention map becomes significantly sharper with lower entropy, allowing nodes to focus on more relevant regions and produce more diverse representations.}
\label{fig:att_vis}
\end{figure}

To mitigate the high-entropy problem commonly observed in standard linear attention mechanisms, we propose a novel attention modulation function that explicitly reduces the row-wise entropy of the attention distribution while maintaining training stability. 
According to the theoretical results established in~\cite{meng2025polaformer}, using a modulation function $ f(x) $ that satisfies $ f'(x) > 0 $ and $ f''(x) > 0 $ to transform the queries and keys, \eg, setting $ \overline \mQ = f(\mQ) $, $ \overline \mK = f(\mK) $, can effectively reduce the Positive Sequence Entropy of the resulting attention matrix during attention computation.

\begin{table*}[t!]
\centering
\renewcommand\tabcolsep{3.8pt}
\resizebox{\textwidth}{!}{
\begin{tabular}{l|l|cccccc|c}
\hline
\textbf{Category} & \textbf{Model} & \textbf{Squirrel} & \textbf{Amazon-Ratings} & \textbf{Roman-Empire} & \textbf{Minesweeper} & \textbf{Questions} & \textbf{Tolokers} & \textbf{Avg. Rank} \\
\hline
\multirow{8}{*}{GNN} 
& GCN  ~\cite{kipf2016semi}     & 38.67 ± 1.84 & 48.70 ± 0.63 & 73.69 ± 0.74 & 89.75 ± 0.52 & 76.09 ± 1.27 & 83.64 ± 0.67 & 11.2 \\
& GAT~\cite{velivckovic2017graph} & 35.62 ± 2.06 & 52.70 ± 0.62 & 88.75 ± 0.41 & 93.91 ± 0.35 & 76.79 ± 0.71 & \textbf{\textcolor{BurntOrange}{83.78 ± 0.43}} & 8.5 \\
& GraphSAGE~\cite{hamilton2017inductive}  & 36.09 ± 1.99 & 53.63 ± 0.39 & 85.74 ± 0.67 & 93.51 ± 0.57 & 76.44 ± 0.62 & 82.43 ± 0.44 & 8.8 \\
& H2GCN ~\cite{zhu2020beyond}     & 35.10 ± 1.15 & 36.47 ± 0.20 & 60.11 ± 0.52 & 89.71 ± 0.31 & 63.59 ± 1.46 & 73.35 ± 1.01 & 16.3 \\
& CPGNN  ~\cite{zhu2021graph}    & 30.04 ± 2.03 & 39.79 ± 0.77 & 63.96 ± 0.62 & 52.03 ± 5.46 & 65.96 ± 1.95 & 73.36 ± 1.01 & 16.1 \\
& GPRGNN ~\cite{chien2020adaptive}   & 38.95 ± 1.99 & 44.88 ± 0.34 & 64.85 ± 0.27 & 86.24 ± 0.61 & 55.48 ± 0.91 & 72.94 ± 0.97 & 14.8 \\
& FSGNN  ~\cite{maurya2022simplifying}    & 35.92 ± 1.32 & 52.74 ± 0.83 & 79.92 ± 0.56 & 90.08 ± 0.70 & 78.86 ± 0.92 & 82.76 ± 0.61 & 9.7 \\
& GloGNN  ~\cite{li2022finding}   & 35.11 ± 1.24 & 36.89 ± 0.14 & 59.63 ± 0.69 & 51.08 ± 1.23 & 65.74 ± 1.19 & 73.39 ± 1.17 & 16.5 \\
\hline
\multirow{3}{*}{Tuned-GNN}
& tuned-GCN ~\cite{luo2024classic}  & \textbf{\textcolor{MidnightBlue}{45.01 ± 1.63}} & 53.80 ± 0.60 & \textbf{\textcolor{BurntOrange}{91.27 ± 0.20}} & \textbf{\textcolor{MidnightBlue}{97.86 ± 0.24}} & \textbf{\textcolor{MidnightBlue}{79.02 ± 0.60}} & -- & \textbf{\textcolor{MidnightBlue}{3.0}} \\
& tuned-SAGE ~\cite{luo2024classic} & 40.78 ± 1.40 & \textbf{\textcolor{BurntOrange}{55.40 ± 0.21}} & 91.06 ± 0.27 & \textbf{\textcolor{BurntOrange}{97.77 ± 0.62}} & 77.21 ± 1.28 & -- & 4.8 \\
& tuned-GAT ~\cite{luo2024classic}  & 41.73 ± 2.07 & \textbf{\textcolor{MidnightBlue}{55.54 ± 0.51}} & 90.63 ± 0.14 & 97.73 ± 0.73 & 77.95 ± 0.51 & -- & 4.6 \\
\hline
\multirow{7}{*}{GT}
& GraphGPS ~\cite{rampavsek2022recipe}   & 39.67 ± 2.84 & 53.10 ± 0.42 & 82.00 ± 0.61 & 90.63 ± 0.67 & 71.73 ± 1.47 & 83.71 ± 0.48 & 9.5 \\
& NAGphormer~\cite{chen2023nagphormer} & -- & 51.26 ± 0.72 & 74.34 ± 0.77 & 84.19 ± 0.66 & 68.17 ± 1.53 & 78.32 ± 0.95 & 13.2 \\
& Exphormer ~\cite{shirzad2023exphormer} & -- & 53.51 ± 0.46 & 89.03 ± 0.37 & 90.74 ± 0.53 & 73.94 ± 1.06 & 83.77 ± 0.78 & 8.2 \\
& NodeFormer ~\cite{wu2022nodeformer} & 38.52 ± 1.57 & 43.86 ± 0.35 & 74.83 ± 0.81  & 87.71 ± 0.69 & 74.27 ± 1.46 & 78.10 ± 1.03 & 12.8 \\
& SGFormer ~\cite{wu2023sgformer}   & 41.80 ± 2.27 & 48.01 ± 0.49 & 79.10 ± 0.32 & 90.89 ± 0.58 & 72.15 ± 1.31 & -- & 10.4 \\
& Polynormer ~\cite{deng2024polynormer} & 40.87 ± 1.96 & 54.81 ± 0.49 & \textbf{\textcolor{MidnightBlue}{92.55 ± 0.37}} & 97.46 ± 0.36 & \textbf{\textcolor{BurntOrange}{78.92 ± 0.89}} & \textbf{\textcolor{MidnightBlue}{85.91 ± 0.74}} & \textbf{\textcolor{BurntOrange}{3.8}} \\
& GQT ~\cite{wang2025learning}  & \textbf{\textcolor{BurntOrange}{42.72 ± 1.69}} & 54.32 ± 0.41 & 90.98 ± 0.24 & 97.36 ± 0.35 & 78.94 ± 0.86 & -- & 4.4 \\
& \textbf{GraphTARIF (Ours)} & \textbf{\textcolor{ForestGreen}{45.58 ± 1.91}} & \textbf{\textcolor{ForestGreen}{55.86 ± 0.42}} & \textbf{\textcolor{ForestGreen}{93.23 ± 0.38}} & \textbf{\textcolor{ForestGreen}{99.03 ± 0.19}} & \textbf{\textcolor{ForestGreen}{79.64 ± 0.71}} & \textbf{\textcolor{ForestGreen}{86.60 ± 0.51}} & \textbf{\textcolor{ForestGreen}{1.0}}  \\
\hline

\end{tabular}
}
\caption{Performance (Accuracy or ROC AUC \% $\pm$ std) on heterophilic datasets with average rank (lower is better). We report ROC AUC for the Minesweeper, Tolokers, and Questions datasets, while Accuracy is reported for Roman-Empire, Amazon-Ratings, and Squirrel. We highlight the top \textbf{\textcolor{ForestGreen}{first}}, \textbf{\textcolor{MidnightBlue}{second}}, and \textbf{\textcolor{BurntOrange}{third}} results per dataset.}
\label{tab:het_performance}
\end{table*}

Previous approaches~\cite{meng2025polaformer,han2023flatten} have explored simple power functions such as $f(x) = x^p$ to sharpen attention distributions. While effective in compressing entropy in the small-value regime, these functions tend to grow rapidly for large $x$, leading to exploding gradients and instability during training.

To overcome these limitations, we propose the following entropy-shaping function:
\begin{equation}
{f(x;p,q) = x \cdot \left(\log(1 + x^p)\right)^q,}
\end{equation}
where the exponents $ p $ and $ q $ are parameterized as $ p = 1 + \alpha \cdot \sigma(w) $ and $ q = 1 + \beta \cdot \sigma(w) $.
with $ \sigma(\cdot) $ denoting the sigmoid function, $ w $ learnable weights, and $ \alpha, \beta > 0 $ as hyperparameters. 
This function possesses several desirable properties for attention modulation, as formally stated in Theorem~\ref{theorem:3}.
The detailed proof is provided in Appendix~\ref{sec:app_proof3}.

This function is applied element-wise to the transformed queries and keys before the attention map is computed. As illustrated in Fig.~\ref{fig:att_vis}, it modulates the magnitude of each feature dimension to generate more concentrated and discriminative attention weights.

\begin{theorem}\label{theorem:3}  
Let $f(x) = x \cdot \left(\log(1 + x^p)\right)^q$ with $p, q > 1$. Then:
\begin{enumerate}
\item $f'(x) > 0 $ and $ f''(x) > 0$ for all $x > 0$, implying reduced attention entropy~\cite{meng2025polaformer}.
\item As $x \to \infty $, $ f'(x) = \mathcal{O}\left((\log x)^q\right)$, which grows slower than power functions (\eg, $ x^p $), promoting smoother gradients and improved training stability.
\end{enumerate}
\end{theorem}

Moreover, our function introduces learnable degrees of freedom through the parameters of both inner ($p$) and outer ($q$) transformations, enabling more fine-grained control over attention sharpness. This enables adaptive attention sharpening where needed, while preserving numerical stability.

\section{Experiments}

\begin{table*}[t!]
\centering
\renewcommand\tabcolsep{7pt}
\resizebox{0.9\textwidth}{!}{
\begin{tabular}{l|l|ccccc|c}
\hline
\textbf{Category} & \textbf{Model} & \textbf{Computer} & \textbf{Photo} & \textbf{CS} & \textbf{Physics} & \textbf{WikiCS} & \textbf{Avg. Rank} \\
\hline
\multirow{9}{*}{GNN} 
& GCN ~\cite{kipf2016semi}& 89.65 ± 0.52 & 92.70 ± 0.20 & 92.92 ± 0.12 & 96.18 ± 0.07 & 77.47 ± 0.85 & 17.2 \\
& GAT ~\cite{velivckovic2017graph}          & 90.78 ± 0.13 & 93.87 ± 0.11 & 93.61 ± 0.14 & 96.17 ± 0.08 & 76.91 ± 0.82 & 16.8 \\
& GraphSAGE ~\cite{hamilton2017inductive}    & 91.20 ± 0.29 & 94.59 ± 0.14 & 93.91 ± 0.13 & 96.49 ± 0.06 & 74.77 ± 0.95 & 14.4 \\
& APPNP~\cite{gasteiger2018predict} & 90.18 ± 0.17 & 94.32 ± 0.14 & 94.49 ± 0.07 & 96.54 ± 0.07 & 78.87 ± 0.11 & 13.4 \\
& GCNII ~\cite{chen2020simple}        & 91.04 ± 0.41 & 94.30 ± 0.20 & 92.22 ± 0.14 & 95.97 ± 0.11 & 78.68 ± 0.55 & 15.8 \\
& GPRGNN~\cite{chien2020adaptive}        & 89.32 ± 0.29 & 94.49 ± 0.14 & 95.13 ± 0.09 & 96.85 ± 0.08 & 78.12 ± 0.23 & 13.4 \\
& GGCN ~\cite{yan2022two}        & 91.81 ± 0.20 & 94.50 ± 0.11 & 95.25 ± 0.05 & 97.07 ± 0.05 & 78.44 ± 0.53 & 10.4 \\
& OrderedGNN ~\cite{song2023ordered}  & 92.03 ± 0.13 & 95.10 ± 0.20 & 95.00 ± 0.10 & 97.00 ± 0.08 & 79.01 ± 0.68 & 8.8 \\
& tGNN  ~\cite{hua2022high}        & 83.40 ± 1.33 & 89.92 ± 0.72 & 92.85 ± 0.48 & 96.24 ± 0.24 & 71.49 ± 1.05 & 19.2 \\
\hline
\multirow{3}{*}{Tuned-GNN} 
& tuned-GCN ~\cite{luo2024classic}    & \textbf{\textcolor{BurntOrange}{93.99 ± 0.12}} & 96.10 ± 0.46 & 96.17 ± 0.06 & \textbf{\textcolor{MidnightBlue}{97.46 ± 0.10}} & 80.30 ± 0.62 & \textbf{\textcolor{MidnightBlue}{3.6}} \\
& tuned-SAGE ~\cite{luo2024classic}   & 93.25 ± 0.14 & \textbf{\textcolor{MidnightBlue}{96.78 ± 0.23}} & \textbf{\textcolor{MidnightBlue}{96.38 ± 0.11}} & 97.19 ± 0.05 & 80.69 ± 0.31 & 4.0 \\
& tuned-GAT ~\cite{luo2024classic}    & \textbf{\textcolor{MidnightBlue}{94.09 ± 0.37}} & 96.60 ± 0.33 & 96.21 ± 0.14 & 97.25 ± 0.06 & \textbf{\textcolor{ForestGreen}{81.07 ± 0.54}} & \textbf{\textcolor{BurntOrange}{3.0}} \\
\hline
\multirow{7}{*}{GT} 
& GraphGPS  ~\cite{rampavsek2022recipe}    & 91.19 ± 0.54 & 95.06 ± 0.13 & 93.93 ± 0.12 & 97.12 ± 0.19 & 78.66 ± 0.49 & 11.4 \\
& NAGphormer  ~\cite{chen2023nagphormer}  & 91.22 ± 0.14 & 95.49 ± 0.11 & 95.75 ± 0.09 & 97.34 ± 0.03 & 77.16 ± 0.72 & 8.6 \\
& Exphormer ~\cite{shirzad2023exphormer}    & 91.47 ± 0.17 & 95.35 ± 0.22 & 94.93 ± 0.01 & 96.89 ± 0.09 & 78.54 ± 0.49 & 10.4 \\
& NodeFormer ~\cite{wu2022nodeformer}   & 86.98 ± 0.62 & 93.46 ± 0.35 & 95.64 ± 0.22 & 96.45 ± 0.28 & 74.73 ± 0.94 & 15.6 \\
& SGFormer ~\cite{wu2023sgformer}     & 91.99 ± 0.70 & 95.10 ± 0.47 & 94.78 ± 0.20 & 96.60 ± 0.18 & 73.46 ± 0.56 & 12.4 \\
& Polynormer ~\cite{deng2024polynormer}   & \textbf{\textcolor{BurntOrange}{93.68 ± 0.21}} & 96.46 ± 0.26 & 95.53 ± 0.16 & 97.27 ± 0.08 & 80.10 ± 0.67 & 5.4 \\
& GQT ~\cite{wang2025learning}   & 93.37 ± 0.44 & 95.73 ± 0.18 & 96.11 ± 0.09 & \textbf{\textcolor{ForestGreen}{97.53 ± 0.06}} & \textbf{\textcolor{BurntOrange}{80.14 ± 0.57}} & 4.4 \\
& \textbf{GraphTARIF (Ours)} & \textbf{\textcolor{ForestGreen}{94.61 ± 0.17}} & \textbf{\textcolor{ForestGreen}{97.03 ± 0.19}} & \textbf{\textcolor{ForestGreen}{96.51 ± 0.11}} & \textbf{\textcolor{BurntOrange}{97.39 ± 0.07}}  & \textbf{\textcolor{MidnightBlue}{80.93 ± 0.57}} & \textbf{\textcolor{ForestGreen}{1.6}} \\
\hline
\end{tabular}
}

\caption{Performance (Accuracy \% $\pm$ std) on homophilic datasets with average rank (lower is better). 
}
\label{tab:hom_performance}
\end{table*}

\subsection{Experimental Setup}

\subsubsection{Datasets}
We evaluate our model on five widely used homophilic graph datasets and six heterophilous graph datasets. The homophilic graphs include Computer, Photo~\cite{mcauley2015image}, CS, Physics~\cite{shchur2018pitfalls}, and WikiCS~\cite{mernyei2020wiki}. The heterophilic graphs in our study are Squirrel~\cite{rozemberczki2021multi}, Roman-Empire, Amazon-ratings, Minesweeper, Tolokers, and Questions~\cite{platonov2023critical}. For these datasets, we adopt the data splitting strategy proposed in~\cite{luo2024classic,deng2024polynormer}. 
In addition, to examine the scalability of our approach, we evaluate it on three large-scale graphs: ogbn-arxiv, ogbn-products, and Pokec. These datasets contain node sizes ranging from 0.16M up to 2.4M, and we follow the standard evaluation settings of~\cite{deng2024polynormer}.

\subsubsection{Baselines} We compare the proposed model against a broad set of baselines covering both classical GNNs and recent Graph Transformer architectures. Specifically, we include: (i) classical GNNs such as GCN, and GraphSAGE. 
; (ii) heterophily-specialized GNNs, such as H2GCN, CPGNN, and TunedGNN, which is a recent framework that enhances traditional GNNs through skip-connections and hyperparameter optimization.
and (iii) Graph Transformer models, such as GraphGPS, Exphormer, and NodeFormer. 
These baselines span a wide range of model paradigms and design philosophies, and collectively represent the current state-of-the-art under both homophilic and heterophilic settings. For baseline methods, we report the results as documented in prior work. For our method, we report the average performance over five independent runs.


\subsection{Main Comparison}

We evaluate the proposed model, GraphTARIF, on a comprehensive set of node classification benchmarks, including both homophilic and heterophilic graph datasets. The results are summarized in Tab.~\ref{tab:het_performance} and Tab.~\ref{tab:hom_performance}, where we report the classification accuracy (\%) or ROC-AUC score, along with the average rank across datasets. We compare against three major categories of baselines:  GNNs, Graph Transformers (GT), and Tuned-GNN.

\subsubsection{Results on Heterophilic Datasets}
As shown in Tab.~\ref{tab:het_performance}, our model consistently achieves the best or comparable performance across all six heterophilic datasets: Squirrel, Amazon-Ratings, Roman-Empire, Minesweeper, Questions, and Tolokers. It ranks first on every dataset where evaluation is available, clearly outperforming both tuned GNN variants and state-of-the-art graph transformers. The overall average rank is 1.0, highlighting the robustness and effectiveness of our approach in handling heterophilic graph structures, where traditional GNNs and attention mechanisms often struggle to generalize.

\subsubsection{Results on Homophilic Datasets}

As shown in Tab.~\ref{tab:hom_performance}, our model achieves the best or comparable performance across all five homophilic datasets. Our method achieves the highest average accuracy on the Computer (94.61\%), Photo (97.03\%), and CS (96.51\%) datasets, and obtains the best average rank of 1.4 among all 19 baseline models.  Considering that these baselines—such as the Tuned-GNN~\cite{luo2024classic} models—have been carefully optimized on these mature benchmarks, the improvements achieved by our model are noteworthy and demonstrate its significant potential.
These results validate that our model not only maintains high expressiveness on classical heterophilic graphs but also generalizes well to classical homophilic graphs.

\begin{table}[t!]
\centering
\resizebox{0.48\textwidth}{!}{
\begin{tabular}{l|ccc}
\hline
\textbf{Model} &  \textbf{ogbn-arxiv} & \textbf{pokec} & \textbf{ogbn-products} \\
\hline
GCN~\cite{kipf2016semi} & 71.74 $\pm$ 0.29 & 75.45 $\pm$ 0.17 & 75.64 $\pm$ 0.21  \\
GAT~\cite{velivckovic2017graph} & 72.01 $\pm$ 0.20 & 72.23 $\pm$ 0.18 & 79.45 $\pm$ 0.59  \\
GPRGNN~\cite{chien2020adaptive} & 71.10 $\pm$ 0.12 & 78.83 $\pm$ 0.06 & 79.76 $\pm$ 0.59  \\
LINKX~\cite{lim2021large} & 66.18 $\pm$ 0.33 & 82.04 $\pm$ 0.07 & 71.59 $\pm$ 0.71\\
\hline
Tuned-GCN~\cite{luo2024classic} & \textbf{\textcolor{MidnightBlue}{73.53 ± 0.12}} & \textbf{\textcolor{MidnightBlue}{86.33 $\pm$ 0.17}} & 82.33 $\pm$ 0.19 \\
Tuned-SAGE~\cite{luo2024classic} & 73.00 ± 0.28 & 85.97 $\pm$ 0.21 & \textbf{\textcolor{ForestGreen}{83.89 $\pm$ 0.36}} \\
Tuned-GAT~\cite{luo2024classic} & 73.30 ± 0.18 & 86.19 $\pm$ 0.23  & 80.99 $\pm$ 0.16 \\
\hline
GraphGPS~\cite{rampavsek2022recipe} & 70.97 $\pm$ 0.41 & \textit{OOM}  & \textit{OOM} \\
NAGphormer~\cite{chen2023nagphormer} & 70.13 $\pm$ 0.55 & 76.59 $\pm$ 0.25 & 73.55 $\pm$ 0.21 \\
Exphormer~\cite{shirzad2023exphormer} & 72.14 $\pm$ 0.28 & \textit{OOM} & \textit{OOM} \\
NodeFormer~\cite{wu2022nodeformer} & 67.19 $\pm$ 0.83 & 71.01 $\pm$ 0.20 & 72.93 $\pm$ 0.13 \\
GOAT~\cite{kong2023goat} & 72.41 $\pm$ 0.40 & 66.37 $\pm$ 0.94 & 82.00 $\pm$ 0.43 \\
SGFormer~\cite{wu2023sgformer} & 72.63 $\pm$ 0.13 & 73.76 $\pm$ 0.24 & 74.16 $\pm$ 0.31 \\
Polynormer~\cite{deng2024polynormer} & \textbf{\textcolor{BurntOrange}{73.46 $\pm$ 0.16}} & 86.10 $\pm$ 0.05 & \textbf{\textcolor{MidnightBlue}{83.82 $\pm$ 0.11}} \\
GQT~\cite{wang2025learning} & 73.14 $\pm$ 0.16 & 83.76 $\pm$ 0.24 & 82.46 $\pm$ 0.17 \\
\hline
\textbf{GraphTARIF (Ours)} & \textbf{\textcolor{ForestGreen}{73.81 $\pm$ 0.11}} & \textbf{\textcolor{ForestGreen}{86.55 $\pm$ 0.07}} & \textbf{\textcolor{BurntOrange}{83.65 $\pm$ 0.27}}\\
\hline
\end{tabular}
}
\caption{Performance (Accuracy \% $\pm$ std) on large-scale datasets. 
\textit{OOM} indicates out of memory. 
}
\label{tab:large}
\end{table}

\subsubsection{Results on Large-scale Graphs}

To further demonstrate the scalability of our approach on large-scale graphs, we evaluate GraphTARIF on three widely used benchmarks: ogbn-arxiv, pokec, and ogbn-products. Among these benchmarks, ogbn-arxiv and ogbn-products are homophilic graphs, while pokec is heterophilic. As shown in Tab.~\ref{tab:large}, GraphTARIF achieves the best performance on ogbn-arxiv and on pokec. On ogbn-products, GraphTARIF also delivers highly competitive results. These results demonstrate that GraphTARIF can effectively scale to large-scale graphs and consistently achieves superior or competitive performance compared to strong baselines.


\begin{table*}[t!]
\centering
\renewcommand\tabcolsep{2.5pt}
\resizebox{1.0\linewidth}{!}{
\begin{tabular}{l|c|c|c|c|cccc}
\hline
\textbf{Model Variant} & $\psi(\mX) \odot \mZ$ & $f(\cdot)$ & Rank-Aug. & Gating Mechanism & \textbf{Minesweeper} & \textbf{Roman-Empire} & \textbf{Computer} & \textbf{Photo} \\
\hline
\textbf{GraphTARIF} (Full)  & $\checkmark$ & $\checkmark$ & $\checkmark$ & $\checkmark$ & \textbf{99.03 ± 0.19} & \textbf{93.23 ± 0.38} & \textbf{94.61 ± 0.17} & \textbf{97.03 ± 0.19} \\ \hline
w/o Post-Modu.  &  & $\checkmark$ & $\checkmark$ & $\checkmark$ & 97.93 ± 0.44 & 90.76 ± 0.55 & 94.24 ± 0.16 & 96.69 ± 0.10 \\
w/o Post-Modu. \& Sharpening Func. &  &  & $\checkmark$ & $\checkmark$ & 94.46 ± 0.89 & 90.44 ± 0.88 & 94.10 ± 0.35 & 96.34 ± 0.25 \\
w/o Gating Mechanism  & $\checkmark$ & $\checkmark$ & $\checkmark$ &  & 98.05 ± 0.44 & 92.23 ± 0.34 & 94.23 ± 0.25 & 96.61 ± 0.27 \\
Vanilla Linear attention  & & & &  & 91.62 ± 0.54 & 89.39 ± 0.42 & 93.75 ± 0.48 & 96.30 ± 0.29 \\
\hline
\end{tabular}
}
\caption{Ablation study of \textsc{GraphTARIF}. 
``$\psi(\mX) \odot \mZ$'' denotes node-wise post-modulation, and ``$f(\cdot)$'' denotes the sharpening function, ``Rank-Aug.'' represents the GAT-based rank augmentation branch.
}
\label{tab:ablation_tarif}
\end{table*}



\subsection{Ablation Study}

To assess the contribution of each component in GraphTARIF, we conduct an ablation study by progressively removing key modules, as summarized in Tab.~\ref{tab:ablation_tarif}. We evaluated four critical components: (1) the node-wise post-modulation $\psi(\mX) \odot \mZ$, which reweights the attention output based on transformed node features; (2) the sharpening function $f(\cdot)$, a learnable log-power transformation that reduces attention entropy; (3) the rank augmentation branch, implemented via a GAT module, increases the rank of the attention map.; and (4) the learnable GAT gate mechanism, which regulates the influence of the GAT branch through a gated scalar. 
We observe that removing any single component leads to a consistent performance drop across all datasets. In particular, removing both post-modulation and the sharpening function further exacerbates the decline, confirming the importance of entropy reduction (third row). Additionally, removing the learnable GAT gate (fourth row) while keeping the GAT branch active also leads to a decline in performance, suggesting that unregulated local attention can interfere with global signal modeling. The full model, which combines all components, consistently achieves the best results across all datasets.
These results collectively validate that each component in GraphTARIF plays a complementary and non-redundant role.

\begin{figure}[t!]
\centering
\includegraphics[width=1.0\linewidth]{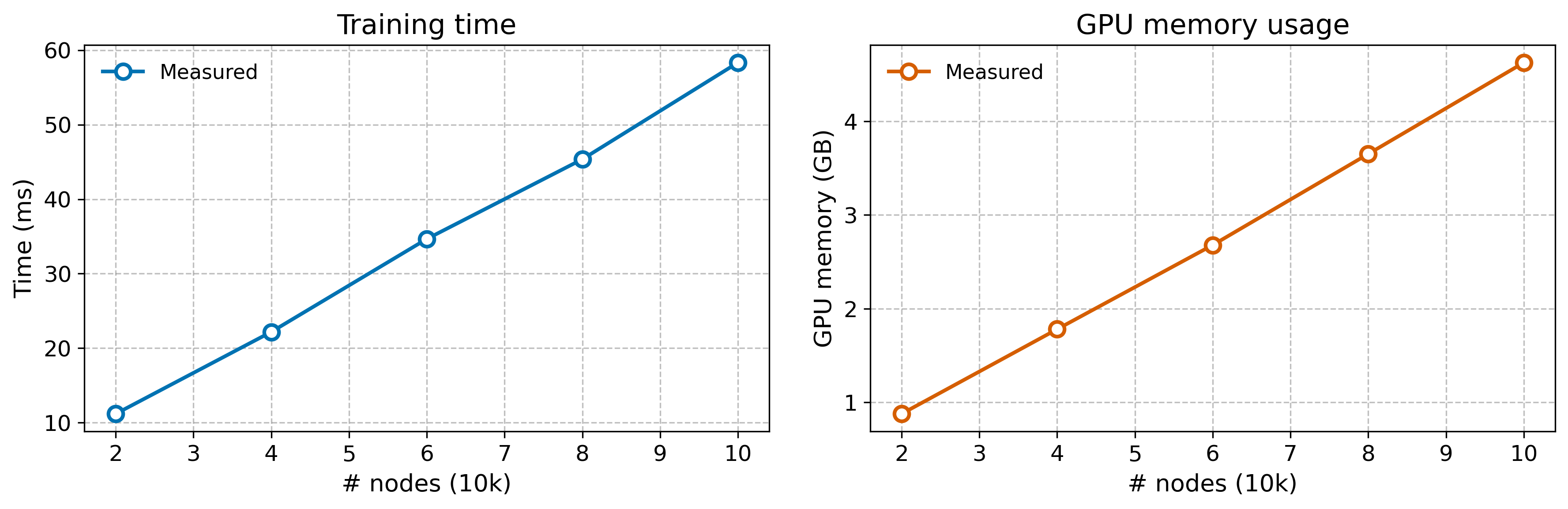}
\caption{Training time and GPU memory usage of GraphTARIF on the pokec dataset. }
\label{fig:time-mem}
\end{figure}

\begin{figure}[t!]
\centering
\includegraphics[width=0.48\textwidth]{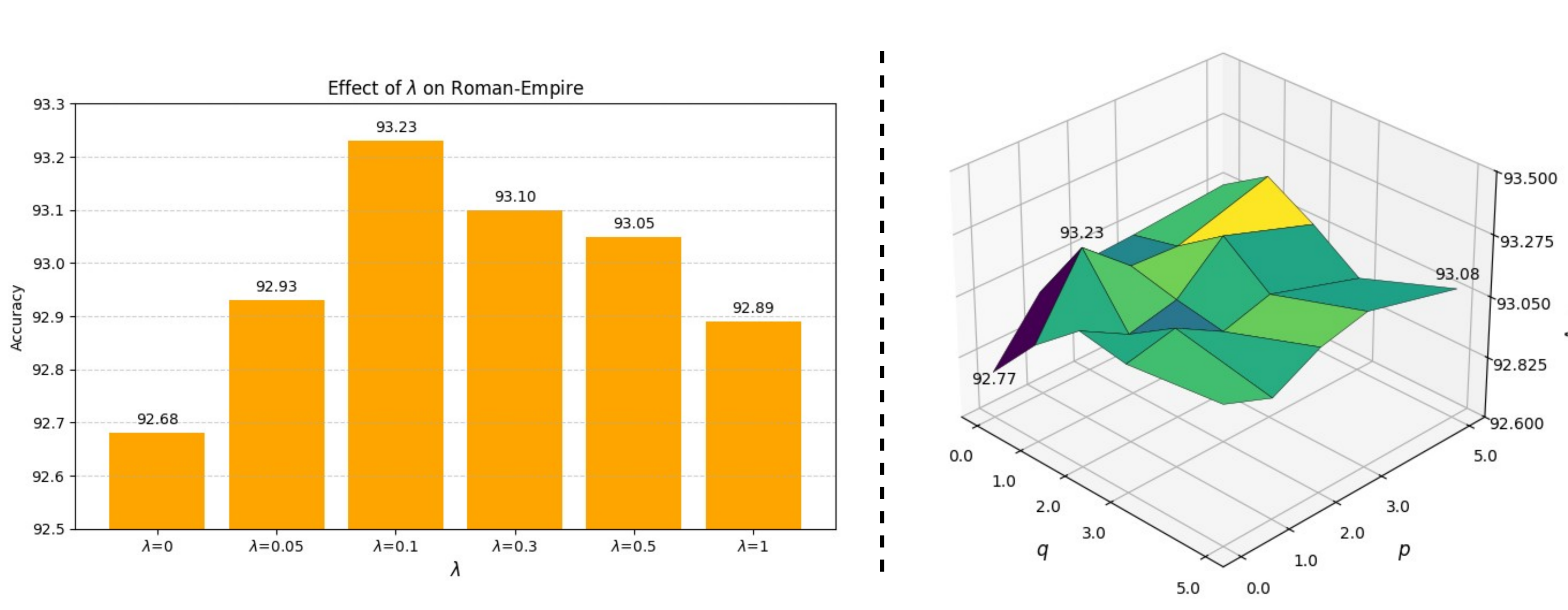} 
\caption{Impact of the hyperparameters $\lambda$ (left), $q$, and $p$ (right) on the Roman-Empire dataset.}
\label{fig:hyp}
\end{figure}


\subsection{Additional Experiments}
\subsubsection{Model Scalability}
This experiment aims to comprehensively evaluate the scalability of GraphTARIF. Specifically, we randomly sampled nodes from the training set, incrementally increasing the sample size from 20,000 to 100,000, and monitored the corresponding training time and GPU memory consumption as the dataset size changed. Fig.~\ref{fig:time-mem} (left) and (right) illustrate the variation in runtime and GPU memory usage, respectively, on the pokec dataset. Evidently, both the runtime and memory consumption of GraphTARIF increase linearly with the growth in graph size, indicating that GraphTARIF exhibits linear time and space complexity. These results clearly demonstrate the scalability of GraphTARIF.
\subsubsection{Hyperparameter Analysis}
These experiments aim to intuitively illustrate the effects of hyperparameter selection. Specifically, we test three key hyperparameters of our method ($\lambda$, $q$, and $p$) on the Roman-Empire dataset. As shown in Fig.~\ref{fig:hyp}, the left panel presents the influence of the coefficient $\lambda$. Evidently, a moderate value ($\lambda = 0.1$) yields the highest accuracy. The right panel of Fig.~\ref{fig:hyp} illustrates the impact of the hyperparameters $p$ and $q$ in the Learnable Log-Power Functions on the Roman-Empire dataset. We observe that the model performance peaks when $p$ and $q$ are set around 1–2. Beyond this range, the performance gradually stabilizes, indicating that moderate values of $p$ and $q$ are sufficient to enhance the model’s expressiveness. Through automated hyperparameter optimization, we further determined that the optimal values of $p$ and $q$ across different datasets typically lie within the range of 1 to 3. At the same time, the model maintains competitive performance even with larger values, further validating the stability of our approach.

\subsubsection{Efficiency Comparison}
As shown in Fig.~\ref{fig:intro11}, compared with recent Graph Transformers, GraphTARIF achieves superior performance while requiring less runtime and memory. 
It attains the best results, and it incurs lower computational costs than Polynormer and Exphormer, demonstrating its effectiveness and efficiency. Moreover, our method achieves strong performance with a relatively small number of parameters, as detailed in Table~\ref{tab:time} of Appendix~\ref{sec:additional_results}.




\section{Conclusion}
In this paper, we introduced GraphTARIF, a hybrid linear Graph Transformer that augments the rank of attention matrices and reduces entropy through a gated local branch and a learnable log-power function. Theoretical analysis and extensive experiments demonstrate that GraphTARIF achieves competitive expressiveness while preserving linear scalability, consistently outperforming strong baselines on both homophilic and heterophilic benchmarks as well as large-scale graphs.

\begin{acks}
This work was supported by the National Natural Science
Foundation of China (92570101) and the Fundamental Research Funds for the Central Universities (226-202500080).
\end{acks}

\bibliographystyle{ACM-Reference-Format}
\balance
\bibliography{sample-base}

\appendix

\section{Datasets and Experimental Details}\label{sec:app_experiments}
Table ~\ref{tab:datasets} summarizes the statistics of all thirteen datasets used in our experiments, along with the corresponding major hyperparameter settings.
We adopt GATConv as the base GNN operator preceding the attention layer, combined with residual connections, layer normalization, and ReLU activation.
The attention module employs a linear attention mechanism with a sigmoid kernel feature map, enabling efficient global information aggregation. After the attention layer, one or two additional GNN layers are stacked to refine node representations and perform classification.
All models are trained in a semi-supervised framework using the PyTorch Geometric (PyG) library. Experiments are conducted on a GPU cluster equipped with NVIDIA RTX 4090, RTX A6000, and L40S devices. Hyperparameters are optimized using Bayesian optimization via Optuna toolkit.

\begin{table*}[t!]
\centering
\renewcommand\tabcolsep{4pt}
\resizebox{1.0\textwidth}{!}{
\begin{tabular}{l|c|r|r|r|r|c|c|c|c|c|c}
\hline
\textbf{Dataset} & \textbf{Type} & \textbf{\# nodes} & \textbf{\# edges} & \textbf{\# Features} & \textbf{Classes} & \textbf{Metric}  & \textbf{GNN Layers} & \textbf{Attention Layers ($L_A$)} & \textbf{$\lambda$} & \textbf{$p$} & \textbf{$q$} \\
\hline
Computer & Homophily & 13,752 & 245,861 & 767 & 10 & Accuracy & 5  & 1 & 0.1 & 2.0 & 1.0 \\
Photo & Homophily & 7,650 & 119,081 & 745 & 8 & Accuracy & 8  & 1 & 0.1 & 2.0 & 2.0 \\
CS & Homophily & 18,333 & 81,894 & 6,805 & 15 & Accuracy & 5  & 3 & 0.1 & 1.0 & 2.0 \\
Physics & Homophily & 34,493 & 247,962 & 8,415 & 5 & Accuracy & 5  & 3 & 0.1 & 3.0 & 2.0 \\
WikiCS & Homophily & 11,701 & 216,123 & 300 & 10 & Accuracy & 5  & 2 & 0.1 & 1.0 & 2.0 \\
\hline
Squirrel & Heterophily & 2,223 & 46,998 & 2,089 & 5 & Accuracy & 4  & 1 & 0.1 & 2.0 & 1.0 \\
tolokers & Heterophily & 11,758 & 519,000 & 10 & 2 & ROC-AUC & 10  & 2 & 0.1 & 3.0 & 2.0  \\
Roman-Empire & Heterophily & 22,662 & 32,927 & 300 & 18 & Accuracy & 10  & 3 & 0.1 & 1.0 & 1.0 \\
Amazon-Ratings & Heterophily & 24,492 & 493,005 & 300 & 5 & Accuracy& 5 & 1 & 0.1 & 1.0 & 1.0 \\
Minesweeper & Heterophily & 10,000 & 39,304 & 7 & 5 & ROC-AUC & 10 & 2 & 0.1 & 2.0 & 1.5\\
Questions & Heterophily & 48,921 & 153,540 & 301 & 2 & ROC-AUC & 8  & 1 & 0.3 & 2.0 & 1.0 \\
\hline
ogbn-arxiv & Homophily (Large graphs) & 169,343 & 1,166,243 & 128 & 40 & Accuracy & 7  & 1 & 0.2 & 3.0 & 0.1 \\
ogbn-products & Homophily (Large graphs) & 2,449,029 & 61,859,140 & 100 & 47 & Accuracy & 7  & 2 & 0.1 & 2.0 & 1.0 \\
pokec & Heterophily (Large graphs) & 1,632,803 & 30,622,564 & 65 & 2 & Accuracy & 7  & 2 & 0.1 & 2.0 & 1.0 \\
\hline
\end{tabular}
}
\caption{Overview of the datasets used for node classification.}
\label{tab:datasets}
\vspace{-1.0em}
\end{table*}

\section{Theoretical Proof}\label{sec:app_theo_proof}

\subsection{Proof for Theorem 1}
\label{sec:app_proof1}


We make the following assumptions and define the symbols used in the theorem:

\begin{itemize}
\item \textbf{Attention Matrix}: Let $\mM \in \mathbb{R}^{n \times n}$ be the attention matrix, where the elements $\mM_{ij} \in [0, 1]$ for all $i, j$, and each row of the matrix is normalized, \eg, for all $i$,
\begin{equation}
\sum_{j=1}^{n} \mM_{ij} = 1.
\end{equation}

\item \textbf{Node Feature Matrix}: Let $\mX \in \mathbb{R}^{n \times d}$ be the node feature matrix, where $n$ is the number of nodes and $d$ is the dimensionality of the node features. There are $K$ classes, and the $i$-th row of $\mX$ is denoted by $\rvx_i$. The set of nodes in class $k$ is denoted as $C_k$, and its size is $n_k = |C_k|$. The node labels are $y_i \in \{1, 2, \dots, K\}$, and the features of each node are drawn from a class-conditional Gaussian distribution: 
\begin{equation}
\mathbf{x}_i \sim \mathcal{N}(\boldsymbol{\mu}_{y_i}, \Sigma),
\end{equation}
where $\boldsymbol{\mu}_{y_i}$ is the mean of class $y_i$, and $\Sigma$ is the shared covariance matrix for all classes.

\item \textbf{Class Centers}: The class centers are assumed to be orthogonal to each other, \eg, for any $i \neq j$,
\begin{equation}
\boldsymbol{\mu}_i^\top \boldsymbol{\mu}_j = 0.
\end{equation}

\item \textbf{Covariance Matrix}: The covariance matrix is assumed to be $\boldsymbol{\Sigma} = \sigma^2 \mI$, where $\mI$ is the identity matrix and $\sigma^2$ is a constant scalar value.
\end{itemize}

With the above assumptions, we now state the complete form of Theorem~\ref{theorem:1}:

 \textbf{Theorem 1}: Let $\mM \in \mathbb{R}^{n \times n}$ be the attention matrix, and $\mX \in \mathbb{R}^{n \times d}$ be the node feature matrix. Under the assumption that the attention matrix $\mM$ is normalized and data-independent, and the node features $\mX$ follow a class-conditional Gaussian distribution, we have the following relationship between the rank of the attention matrix and the expected value of the trace of the between-class scatter matrix:
\begin{equation}
\mathbb{E}[\text{tr}(\mS_B)] \leq C \cdot r,
\end{equation}
where $r = \text{rank}(\mM)$ is the rank of the attention matrix, and $C$ is a constant.

\begin{proof}
We first consider the feature transformation $\mX' = \mM \mX$.  
Each transformed feature $\rvx_i' = \sum_j \mM_{ij}\rvx_j$ is still Gaussian since $\rvx_j \sim \mathcal{N}(\boldsymbol{\mu}_{y_j},\Sigma)$.  
For class $k$, the empirical center after transformation can be written as
\[
\boldsymbol{\mu}_k' = (\boldsymbol{\alpha}^{(k)})^\top \mX, 
\quad \text{with } \alpha_j^{(k)}=\tfrac{1}{n_k}\sum_{i\in C_k}\mM_{ij},
\]
and the global mean as $\boldsymbol{\mu}'=\boldsymbol{\gamma}^\top \mX$, where $\gamma_j=\tfrac{1}{n}\sum_{i=1}^n\mM_{ij}$.

The between-class scatter becomes
\[
\mS_B = \frac{1}{K}\sum_{k=1}^K(\boldsymbol{\mu}_k'-\boldsymbol{\mu}')(\boldsymbol{\mu}_k'-\boldsymbol{\mu}')^\top
= \mX^\top \mA \mX,
\]
where
\[
\mA = \frac{1}{K}\sum_{k=1}^K (\boldsymbol{\alpha}^{(k)}-\boldsymbol{\gamma})(\boldsymbol{\alpha}^{(k)}-\boldsymbol{\gamma})^\top.
\]
Hence
\[
\mathbb{E}[\operatorname{tr}(\mS_B)] = \operatorname{tr}(\mA\,\mathbb{E}[\mX\mX^\top]).
\]

We decompose $\mX=\mY+\mE$, with $\mY$ the deterministic class means and $\mE$ Gaussian noise.  
This gives
\[
\mathbb{E}[\mX\mX^\top]=\mY\mY^\top+d\sigma^2\mI_n,
\]
so that
\[
\mathbb{E}[\operatorname{tr}(\mS_B)] = \operatorname{tr}(\mA\mY\mY^\top)+d\sigma^2\operatorname{tr}(\mA).
\]

Two auxiliary bounds are needed:

\textbf{Lemma 1.} $\|\boldsymbol{\alpha}^{(k)}-\boldsymbol{\gamma}\|\le \sqrt{n}$, since both are convex combinations of $\{M_{ij}\}$.  

\textbf{Lemma 2.} From Lemma 1,
\[
\|\mA\|_F \le \tfrac{n}{\sqrt{K}}, \quad 
\lambda_{\max}(\mA)\le \|\mA\|_F \le \tfrac{n}{\sqrt{K}}.
\]

Thus $\operatorname{tr}(\mA)\le r\,\lambda_{\max}(\mA)\le \tfrac{nr}{\sqrt{K}}$, with $r=\operatorname{rank}(\mA)\le \operatorname{rank}(\mM)$.

Finally, since both $\mY\mY^\top$ and $\mA$ are PSD,
\[
\operatorname{tr}(\mA\mY\mY^\top)\le \lambda_{\max}(\mY\mY^\top)\operatorname{tr}(\mA).
\]
Combining the two terms yields
\[
\mathbb{E}[\operatorname{tr}(\mS_B)] \le \big(\lambda_{\max}(\mY\mY^\top)+d\sigma^2\big)\cdot \tfrac{nr}{\sqrt{K}}.
\]
Let $C=\tfrac{n}{\sqrt{K}}(\lambda_{\max}(\mY\mY^\top)+d\sigma^2)$, independent of $\mM$, we conclude
\[
\mathbb{E}[\operatorname{tr}(\mS_B)] \le C\cdot \operatorname{rank}(\mM).
\]
\end{proof}

Although Theorem~\ref{theorem:1} provides a upper bound on the inter-group variance, it still offers valuable insight in practice. Real-world data often exhibit complex, multi-modal, or manifold-structured distributions rather than simple class-wise Gaussian clusters. In such cases, a higher-rank attention matrix is still necessary to retain discriminative features across these intricate structures.

\subsection{Proof for Theorem 2}
\label{sec:app_proof2}
\paragraph{Notation.}
We write $\mA \preceq \mB$ to indicate that $\mB - \mA$ is positive semidefinite (PSD), that is, for all $\mathbf{v} \in \mathbb{R}^n$,
\begin{equation}
\mathbf{v}^\top (\mB - \mA) \mathbf{v} \ge 0.
\end{equation}
Equivalently, this means all eigenvalues of $\mB - \mA$ are nonnegative. When $\mA \prec \mB$, the difference is positive definite (PD).

\textbf{Theorem 2}:
Let $\mM_1 = \mP \mM_2$ with $\mP \in \mathbb{R}^{n \times n}$. Since $\mP$ is a smoothing operator, it is reasonable to assume that $\mP$ satisfies $\mP^\top \mP \preceq \mI$. Moreover, assume that $\operatorname{rank}(\mM_1) < \operatorname{rank}(\mM_2)$. Let the input feature matrix be $\mX = \mY + \mE$, where:
\begin{itemize}
\item $\mY \in \mathbb{R}^{n \times d}$ is the matrix of class centers with $Y Y^\top \succ 0$;
\item $\mE \sim \mathcal{N}(0, \sigma^2 I)$ is isotropic Gaussian noise.
\end{itemize}
Then the expected inter-class variance after applying $\mM_1$ is strictly smaller than that of $\mM_2$:
\begin{equation}
\mathbb{E}[\operatorname{tr}(\mS_B(\mM_1))] < \mathbb{E}[\operatorname{tr}(\mS_B(\mM_2))].
\end{equation}

\begin{proof}
For any attention matrix $\mM \in \mathbb{R}^{n \times n}$, define
\[
\boldsymbol{\alpha}^{(k)} = \tfrac{1}{n_k} \sum_{i \in C_k} \mM_{i,:}, \quad
\boldsymbol{\gamma} = \tfrac{1}{n} \sum_{i=1}^n \mM_{i,:},
\]
and
\[
\mA = \tfrac{1}{K} \sum_{k=1}^K (\boldsymbol{\alpha}^{(k)}-\boldsymbol{\gamma})(\boldsymbol{\alpha}^{(k)}-\boldsymbol{\gamma})^\top.
\]
Under $\mM$, the between-class scatter is
\[
\mS_B = \mX^\top \mA \mX, \quad 
\mathbb{E}[\operatorname{tr}(\mS_B)] = \operatorname{tr}(\mA(\mY\mY^\top+d\sigma^2\mI)).
\]

Now consider $\mM_1=\mP\mM_2$, inducing $\mA_1=\mP\mA_2\mP^\top$. Then
\[
\mathbb{E}[\operatorname{tr}(\mS_B(\mM_1))] 
= \operatorname{tr}(\mP\mA_2\mP^\top \mY\mY^\top) 
+ d\sigma^2 \operatorname{tr}(\mP\mA_2\mP^\top).
\]

Let $\mQ=\mP^\top\mP \preceq \mI$. For the signal term,
\[
\operatorname{tr}(\mP\mA_2\mP^\top \mY\mY^\top) 
= \operatorname{tr}(\mA_2 \mP^\top \mY\mY^\top \mP)
< \operatorname{tr}(\mA_2 \mY\mY^\top),
\]
since $\mY\mY^\top \succ 0$ and $\mQ \prec \mI$.  
For the noise term,
\[
\operatorname{tr}(\mP\mA_2\mP^\top) = \operatorname{tr}(\mA_2\mQ) < \operatorname{tr}(\mA_2).
\]

Thus both terms under $\mM_1$ are strictly smaller than under $\mM_2$, giving
\[
\mathbb{E}[\operatorname{tr}(\mS_B(\mM_1))] < \mathbb{E}[\operatorname{tr}(\mS_B(\mM_2))].
\]
\end{proof}


\subsection{Proof for Theorem 3}
\label{sec:app_proof3}
\textbf{Theorem 3}:
Let $ f(x) = x \cdot \left(\log(1 + x^p)\right)^q $, where $ p > 1 $, $ q > 1 $. Then:
\begin{enumerate}
\item $ f'(x) > 0 $ and $ f''(x) > 0 $ for all $ x > 0 $;
\item As $ x \to \infty $, it holds that
\begin{equation}
f'(x) = \mathcal{O}((\log x)^q).
\end{equation}
\end{enumerate}

\begin{proof}
We prove two claims.

\paragraph{(1) Convexity.}  
Let $f(x)=x(\log(1+x^p))^q$ with $L(x)=\log(1+x^p)$. Then
\[
f'(x) = L(x)^q + xqL(x)^{q-1}\frac{p x^{p-1}}{1+x^p} > 0 \quad (x>0).
\]
Differentiating again and writing $f''(x)=A'(x)+B'(x)$ with  
$A(x)=L(x)^q$, $B(x)=xqL(x)^{q-1}L'(x)$, one obtains
\[
B'(x)=pq\Big[L(x)^{q-1}(L'(x)+xL''(x))+(q-1)xL(x)^{q-2}(L'(x))^2\Big].
\]
A direct calculation shows $L'(x)+xL''(x)=\tfrac{p^2x^{p-1}}{(1+x^p)^2}>0$.  
Since every term is nonnegative, we have $B'(x)>0$ and $A'(x)>0$, hence
\[
f''(x) > 0 \quad \text{for all } x>0.
\]

\paragraph{(2) Asymptotic growth.}  
As $x\to\infty$, $\log(1+x^p)\sim p\log x$, and
\[
\frac{px^{p-1}}{1+x^p}\sim \frac{p}{x}, \qquad x\cdot \frac{p}{x}=p.
\]
Thus
\[
f'(x) \sim (p\log x)^q + qp(p\log x)^{q-1} = \mathcal{O}((\log x)^q).
\]

Combining (1) and (2) proves the claim.
\end{proof}

\begin{figure}[htbp]
\centering
\includegraphics[width=0.95\linewidth]{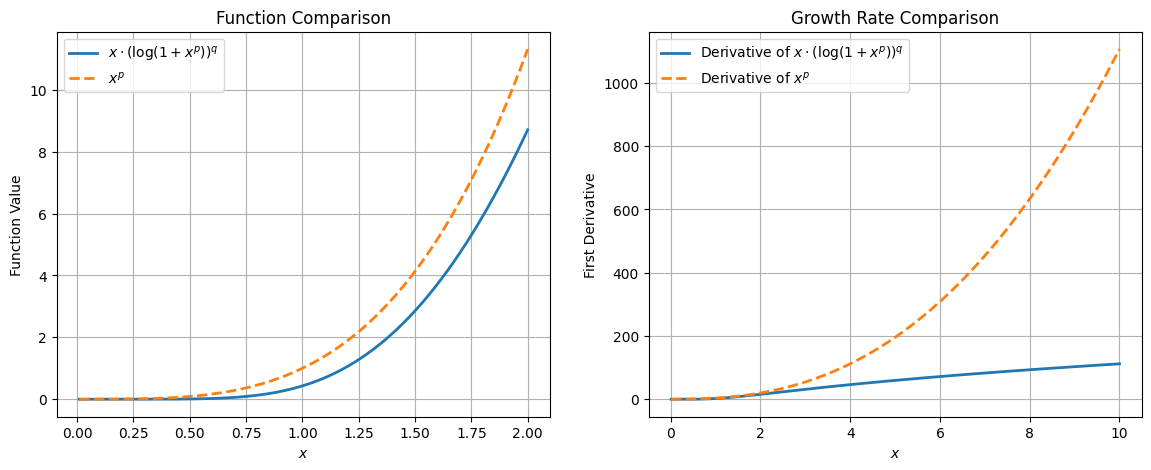}
\vspace{-1.5em}
\caption{Comparison between $x \cdot (\log(1 + x^p))^q$ and $x^p$. Left: function values; Right: growth rates.}
\label{fig:comparison_combined}
\end{figure}

The proposed low-power function $ f(x; p, q) = x \cdot \left(\log(1 + x^p)\right)^q $, similar to the power function, amplifies differences in input values and thus reduces the entropy of the attention distribution, as shown in Fig.~\ref{fig:comparison_combined} (left). However, as illustrated in Fig.~\ref{fig:comparison_combined} (right), its derivative grows significantly more slowly when $x$ is large. This property effectively mitigates the risk of exploding gradients during backpropagation and ensures more stable training.

\subsection{Theoretical Analysis of Node-wise Post-modulation Effects on Output Features}
\label{sec:app_proof4}
In our framework, the node-wise post-modulation operation (Eq.~\ref{post-m}) performs an element-wise transformation on the attention outputs. Prior work~\cite{fan2024breaking} demonstrates that such post-modulation can increase the rank of the feature matrix, thereby enhancing its representational capacity. Complementing this view, we theoretically show that post-modulation also \emph{reduces the entropy} of node features—sharpening their distribution while preserving diversity, which ultimately benefits downstream classification.

\textbf{Theorem 4.}
Let $\mX \in \mathbb{R}^{n \times d}$ denote node features, and $\mM \in \mathbb{R}^{n \times n}$ be an attention matrix emphasizing same-class nodes. Define
\begin{equation}
\mY_1 = \mM\mX, \quad \mY_2 = (\mM\mX) \odot \mX.
\end{equation}
Then for each feature dimension $k$,
\begin{equation}
\mathrm{PSE}(\mY_2^{(:,k)}) < \mathrm{PSE}(\mY_1^{(:,k)}),
\end{equation}
where $\mathrm{PSE}(\cdot)$ denotes the Positive Sequence Entropy.

\begin{proof}
We analyze one feature dimension $\mathbf{x}=(x_1,\dots,x_n)$.  
For two nodes $x=(a,b)$ with $a>b>0$, define $c=a/b>1$.  
The normalized entropy is
\[
\mathrm{PSE}(a,b) = \log(c+1)-\tfrac{c}{c+1}\log c.
\]
Because $\mM$ assigns higher weights to same-class nodes, $(\mM\mathbf{x})_1>(\mM\mathbf{x})_2$ whenever $x_1>x_2$.  
After post-modulation $\mathbf{y}=\mM\mathbf{x}\odot\mathbf{x}$,
\[
\frac{y_1}{y_2}=\frac{(\mM\mathbf{x})_1x_1}{(\mM\mathbf{x})_2x_2}>\frac{x_1}{x_2}=c>1.
\]
Let $h(x)=\log(x+1)-\frac{x}{x+1}\log x$, with derivative $h'(x)=-\frac{\log x}{(x+1)^2}<0$ for $x>1$.  
Since $h(x)$ is strictly decreasing, enlarging the ratio $y_1/y_2$ decreases entropy:
\[
\mathrm{PSE}(y_1,y_2)<\mathrm{PSE}(x_1,x_2).
\]
The result extends to higher dimensions by coordinate-wise analysis, proving that node-wise post-modulation reduces feature entropy.
\end{proof}

\section{Additional Results}
\label{sec:additional_results}
\begin{figure}[htbp]
\centering
\begin{subfigure}[t]{0.38\linewidth}
\centering
\includegraphics[width=\linewidth]{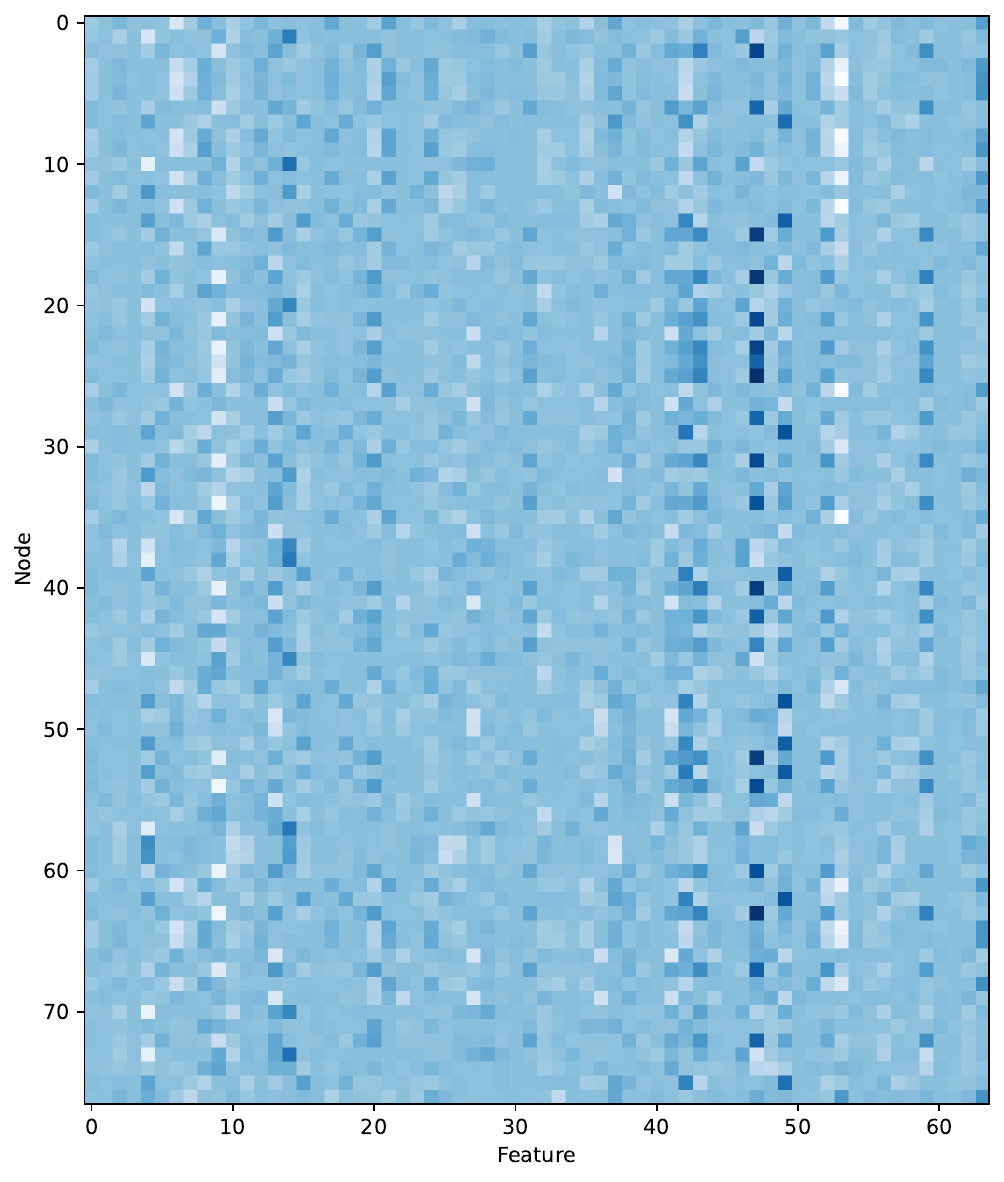}
\vspace{-1.5em}
\caption*{{\centering \scriptsize (a) GraphTARIF, Entropy=3.77}}
\end{subfigure}
\hfill
\begin{subfigure}[t]{0.38\linewidth}
\centering
\includegraphics[width=\linewidth]{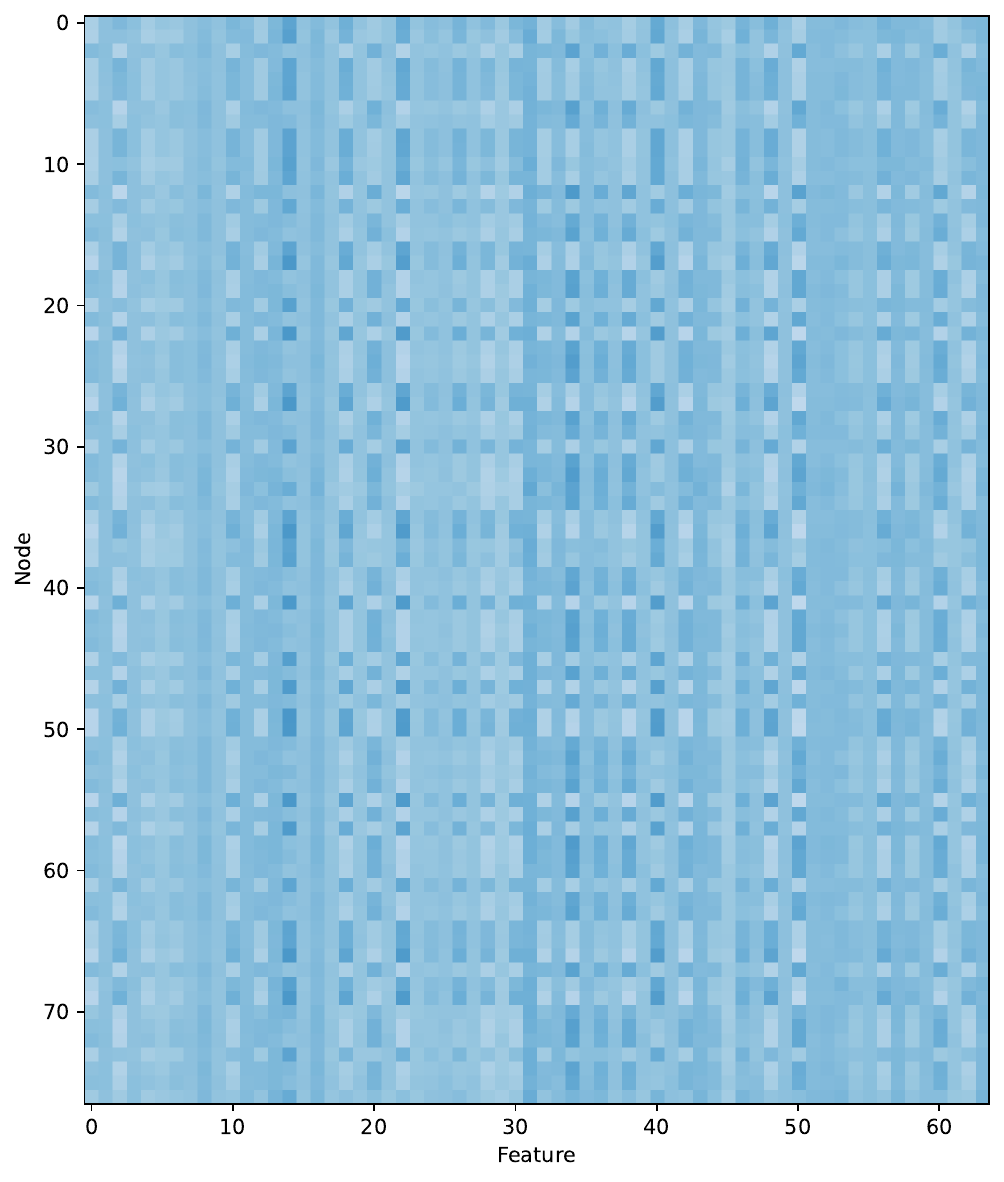}
\vspace{-1.5em}
\caption*{{\centering \scriptsize (b) Linear Attention, Entropy=3.97}}
\end{subfigure}
\caption{Visualization of the output features on the Photo dataset with 
entropy analysis.}
\label{fig:lambda-xrank}
\end{figure}

\begin{figure}[htbp]
\centering
\includegraphics[width=0.45\textwidth]{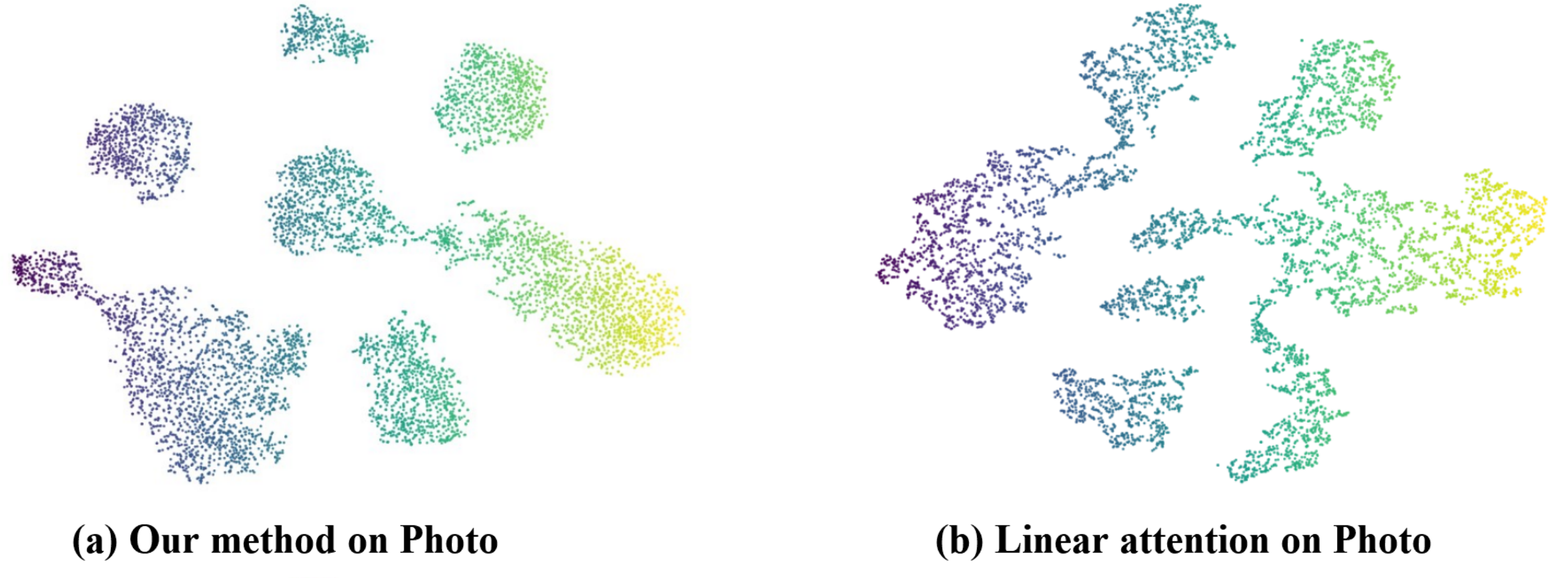}  
\caption{t-SNE visualizations of output features.}
\label{fig:lambda-tsne}
\end{figure}

Tab.~\ref{tab:time} and Fig.~\ref{fig:lambda-xrank}--\ref{fig:lambda-tsne} provide additional results validating the effectiveness and efficiency of GraphTARIF. As shown in Tab.~\ref{tab:time}, GraphTARIF achieves superior performance on two heterophilic graphs with substantially fewer parameters than Polynormer and Exphormer. Fig.~\ref{fig:lambda-xrank} visualizes the first-layer attention features on the Photo dataset, where our method produces representations with higher rank and lower entropy. The t-SNE~\cite{maaten2008visualizing} projection in Fig.~\ref{fig:lambda-tsne} further reveals clearer decision boundaries and greater inter-class separability.

\begin{table}[t!]
\centering
\resizebox{\columnwidth}{!}{
\begin{tabular}{l|cc|cc}
\hline
\multirow{2}{*}{\textbf{Model}} 
& \multicolumn{2}{c|}{\textbf{Roman-Empire}} 
& \multicolumn{2}{c}{\textbf{Questions}} \\
\cline{2-5}
& \textbf{ROC-AUC} & \textbf{Params} 
& \textbf{Accuracy} & \textbf{Params} \\
\hline
Exphormer~\cite{shirzad2023exphormer} 
& 88.42$\pm$0.41 & 15.9M 
& 73.94$\pm$1.06 & 10.7M \\
Polynormer~\cite{deng2024polynormer} 
& \underline{92.55$\pm$0.37} & \underline{9.91M} 
& \underline{78.92$\pm$0.89} & \underline{6.74M} \\
\hline
\textbf{GraphTARIF (Ours)} 
& \textbf{93.23$\pm$0.38} & \textbf{4.07M} 
& \textbf{79.64$\pm$0.71} & \textbf{5.17M} \\
\hline
\end{tabular}
}
\caption{
ROC-AUC or Accuracy (\%) and parameter counts (M) on two heterophilic graph datasets. 
}
\label{tab:time}
\vspace{-2.0em}
\end{table}


\end{document}